\documentclass{article}
\usepackage{semantok,times}

\usepackage{amsmath,amsfonts,amssymb,bm}

\newcommand{\captiona}{{\em (a)}}
\newcommand{\captionb}{{\em (b)}}

\newcommand{\resultlead}[1]{{\fontencoding{OT1}\fontfamily{cmr}\fontseries{bx}\fontshape{n}\selectfont #1}\space}
\newcommand{\resultblock}[1]{\par\vspace{0.35ex}\noindent\resultlead{#1}}

\def\eqref#1{equation~\ref{#1}}

\def\1{\bm{1}}

\DeclareMathAlphabet{\mathsfit}{\encodingdefault}{\sfdefault}{m}{sl}
\SetMathAlphabet{\mathsfit}{bold}{\encodingdefault}{\sfdefault}{bx}{n}

\usepackage{hyperref}
\usepackage{cleveref}
\usepackage{url}
\usepackage{graphicx}
\usepackage{float}
\usepackage{wrapfig}
\usepackage{capt-of}
\usepackage{booktabs}
\usepackage{xcolor}
\usepackage{array}
\usepackage{multirow}
\usepackage{enumitem}
\usepackage{tikz}
\definecolor{fidcol}{HTML}{164279}
\definecolor{semcol}{HTML}{1B4A1D}
\definecolor{stcol}{HTML}{7950B3}
\definecolor{vftcol}{HTML}{D9713A}
\newcommand{\ours}{SemanTok}
\newcommand{\vft}{VideoFlexTok}
\definecolor{find1}{HTML}{9E3429}
\definecolor{find2}{HTML}{705400}
\definecolor{find3}{HTML}{00626B}
\definecolor{find4}{HTML}{AB1456}
\definecolor{find5}{HTML}{5B2C8C}
\definecolor{find6}{HTML}{9A4200}
\definecolor{find7}{HTML}{595959}

\DeclareRobustCommand{\fid}[1]{\textcolor{fidcol}{#1}}
\DeclareRobustCommand{\sem}[1]{\textcolor{semcol}{#1}}
\newcommand{\gFVD}{gFVD}
\newcommand{\gFID}{gFID}
\newcommand{\rFVD}{rFVD}

\newcommand{\PSNR}{PSNR}

\newcommand{\ViCLIP}{ViCLIP}
\newcommand{\ClipV}{ClipV}
\newcommand{\classacc}{class accuracy}
\newcommand{\classaccadj}{class-accuracy}
\newcommand{\decalign}{decoder semantic alignment}
\DeclareRobustCommand{\fbadge}[1]{%
  \tikz[baseline={([yshift=-0.31em]n.center)}]\node[circle,fill=find#1,text=white,inner sep=0pt,
    minimum size=0.90em,font=\sffamily\fontsize{5.5}{5.5}\selectfont\bfseries](n){#1};}
\newcommand{\findinghead}[2]{%
  \par\vspace{0.6ex}\noindent\fbadge{#1}\,%
  \resultlead{\textcolor{find#1}{#2}}}
\newenvironment{findings}{%
  \begin{itemize}[leftmargin=1.6em,labelsep=0.45em,itemsep=4pt,topsep=4pt,parsep=0pt]}%
  {\end{itemize}}
\newcommand{\finding}[2]{\item[\fbadge{#1}]{\fontencoding{OT1}\fontfamily{cmr}\fontseries{bx}\fontshape{n}\selectfont\textcolor{find#1}{#2}}}

\graphicspath{{figures/}}
\newif\ifincludewan
\includewanfalse

\title{SemanTok: Predictable Semantic Tokens for\\ Efficient Autoregressive Video Generation}
\newcommand{\projecturl}{\url{https://semantoken.github.io}}
\author{%
Mikhail Dereviannykh$^{1,2}$\quad
Vikram Voleti$^{1}$\quad
Simon Donn\'e$^{1}$\\
\bfseries Mallikarjun Byrasandra Ramalinga Reddy$^{1}$\quad
Shimon Vainer$^{1}$\quad
Mark Boss$^{1}$\\[3pt]
{\normalfont $^{1}$Stability AI\quad $^{2}$Karlsruhe Institut f\"ur Technologie}\\[2pt]
{\normalfont\small Project page with videos: \projecturl}}

\begin{document}
\maketitle

\begin{abstract}
Recent video-based world models pair the scalability of autoregressive (AR) prediction with the visual quality of diffusion models.
The choice of scene tokenizer is paramount for the optimal performance of each of these, both in terms of fidelity and semantics. Flexible-length, coarse-to-fine tokenizers yield exactly that: the first coarse tokens carry the clip's global semantics while later tokens further specify details.
Existing flexible tokenizers only apply a representation-alignment (REPA) loss on early decoder hidden states, a target the decoder can partly meet from its noised input instead.
We introduce \emph{SemanTok}, a flexible video tokenizer that feeds frozen DINO features into its encoder and adds lightweight heads that reconstruct them from each retained token prefix alone.
\ours{} achieves high semantic alignment and video fidelity at every AR model size: a 201M \ours{} AR model matches or beats a \vft{} AR model $3.4\times$ its size, and larger \ours{} AR models further improve fidelity.
It keeps semantic alignment on out-of-distribution classes and gives the decoder higher semantic alignment at every noise level, including pure noise.
It performs well in both reconstruction and generation, and its short token prefixes are cheaper to predict and give better generation fidelity, with pixel detail deferred to later tokens.

\end{abstract}

\begin{figure*}[!h] 
\centering
\includegraphics[width=\textwidth]{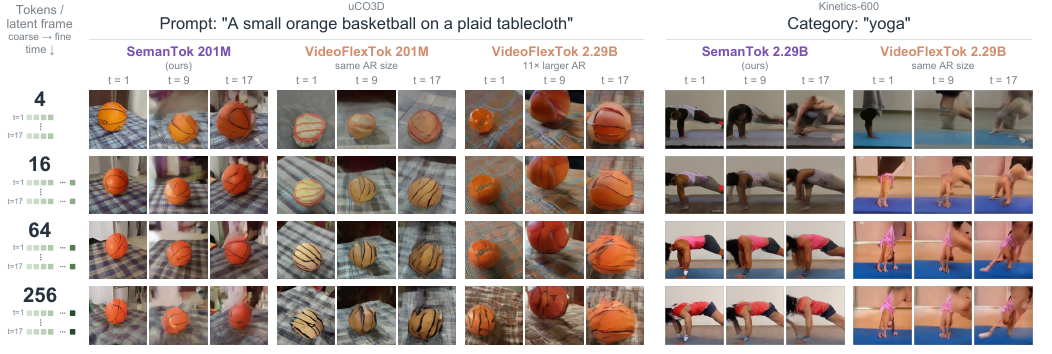}
\vspace{-\parskip}
\vspace{-2.0\abovecaptionskip}
\caption{From a prompt with ``\ldots orange basketball\ldots'' (left,
uCO3D), our 
\ours{} tokenizer lets a 201M autoregressive (AR) model
generate a video that keeps the ball's shape and appearance through the
orbit with a budget of only $k{=}4$ tokens per latent frame.
VideoFlexTok: at the same model's size the ball
is semantically misaligned, and even an $11\times$ larger AR model (2.29B) leaves
its shape unstable up to $k{=}64$. So the smaller, faster model on 
SemanTok
tokens matches or beats the larger one of \vft{}.
Right (Kinetics-600, class ``yoga''): both tokenizers at the same 2.29B AR
size, i.e., equal cost.
\ours{}
keeps a complex body motion stable from $k{=}16$, and larger $k$
refines its appearance and motion. Meanwhile 
\vft{}
changes the scene
between $k{=}4$ and $k{=}16$ and poorly simulates body motion.
Both \vft{} and \ours{} are variable-length: the AR model generates tokens coarse
to fine, and any budget $k$ decodes to a video.
\ours{}
prioritizes semantic content in its first tokens, so a small $k$
already fixes what the clip contains.
Pipeline in \cref{fig:overview}; videos on the project page.}
\label{fig:teaser}
\end{figure*}

\section{Introduction}
\label{sec:intro}

Video clips vary widely in complexity: a static shot has few entities and
little motion, while a car chase has changing viewpoints and detail. To
make such clips tractable, video models compress them into learned tokens
that an autoregressive (AR) model predicts, as a language model predicts
text, and a diffusion decoder renders into
pixels~\citep{yan2021videogpt,yu2023language}. These tokens must
therefore be both easy to predict and informative enough to render. Yet
standard video tokenizers map every clip to the same fixed-size
spatiotemporal grid~\citep{nvidia2025cosmos,tang2024vidtok,yu2023language}.
Because prediction cost grows with sequence length, simple and complex
    clips then get the same representational budget and the same compute.

\begin{figure}[t]
\centering
\includegraphics[width=\textwidth]{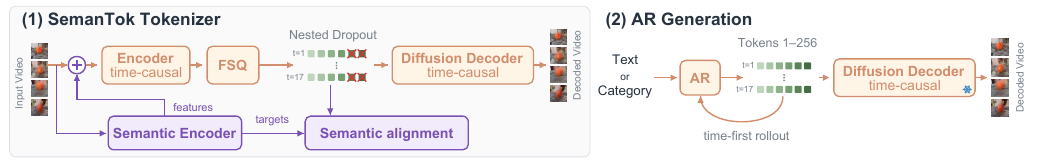}
\vspace{-\parskip}
\vspace{-2.0\abovecaptionskip}
\caption{Overview.
(1)~Tokenizer training: encoder, FSQ, and diffusion decoder are trained
jointly, with nested dropout so that the decoder can reconstruct the clip
from any token prefix.
\ours{} keeps this \vft{} recipe (\textcolor[HTML]{F1A17F}{orange}) and adds semantic supervision from a frozen teacher: its features enter the encoder, and every retained prefix is trained to predict them (\textcolor[HTML]{7950B3}{purple}; details in \cref{fig:arch}).
(2)~AR training and generation: with the tokenizer frozen, an AR model
learns to predict its tokens in time-first, coarse-to-fine order, and the
decoder trained in~(1) renders any generated prefix as video.}
\label{fig:overview}
\end{figure}

Flexible-length, coarse-to-fine
tokenizers~\citep{flextok2025,videoflextok} replace the fixed grid with a
linear sequence that can be meaningfully cropped to various lengths. At
inference time, the application, not the model, selects a budget $k$ and
the autoregressive model predicts only up to that prefix length. Each
prefix should support a high-quality rendering, with additional tokens
only increasing its specificity. We argue that this approach should aim
to carry a clip's semantics in the coarse prefix, so that a small AR model can
settle on what the video should contain, before a larger model spends
capacity on fixing appearance details.
Existing flexible tokenizers use representation alignment
(REPA)~\citep{yu2025repa} at an early decoder layer, showing
that this improves downstream fidelity and semantic alignment~\citep{videoflextok}; this is mostly inspired by image
tokenizers~\citep{zhu2024digit,chen2025maetok,yao2025vavae}.
But that decoder also sees the noised latent, which at low noise
meets the target almost regardless of the tokens
(\cref{fig:dino-probe-main}).

For this, we propose \ours{}, a flexible video tokenizer designed to
prioritize semantic content early in an AR-model rollout (\cref{fig:teaser};
pipeline in \cref{fig:overview}).
We keep the original decoder REPA loss and the coarse-to-fine recipe.
Frozen DINOv2
features~\citep{oquab2023dinov2} enter the encoder alongside the video
latents. Auxiliary heads then read only the retained token prefix and reconstruct the DINO features.
We call these two uses of the teacher \emph{semantic supervision}. No loss assigns information to particular tokens; nested dropout concentrates the most relevant information in the earliest ones.
The result is a variable-length semantic code: every prefix, at any budget, carries the teacher's view of the clip.

The same prefix that these pathways make more semantic is also easier for an AR model to predict. At the same length, the AR model then needs fewer bits to code it: \ours{} keeps what the clip contains early and defers part of the pixel detail the AR model could not predict, a prefix-level form of
the compression--generation trade-off~\citep{ramanujan2024crt,wang2024larp}.

We evaluate \ours{} (\cref{sec:method}) on \fid{fidelity} and \sem{semantic alignment} in the generation
and ground-truth-token reconstruction settings for video (\cref{sec:setup}).
We score seven AR sizes (49M--2.29B) across token budgets, on
class-conditioned (Kinetics-600) and text-conditioned (uCO3D).
Seven findings emerge (\cref{sec:results}):
\begin{findings}
\finding{1}{SemanTok exhibits high semantic alignment and video fidelity at every AR model size.} A 201M \ours{}
AR model matches or beats a \vft{} AR model $3.4\times$ its size
(\cref{fig:teaser,fig:scaling,fig:ar-size,fig:gen-qual-uco3d}; see exception \fbadge{7}).
\finding{2}{Increasing SemanTok model size further improves fidelity.}
The smaller \ours{} AR model exceeds the larger \vft{} AR model in
Kinetics-600 \classacc{} (semantic alignment). Increasing model size further improves fidelity (\gFVD{}) (\cref{fig:ar-size,fig:trainD,fig:guitar-seated,fig:gen-qual-egg}).
\finding{3}{SemanTok is able to maintain semantic alignment over out-of-distribution classes.} Both tokenizer reconstruction and AR generation show the generalization capability of SemanTok on held-out classes (\cref{tab:catacc,fig:recon-idood,fig:gen-qual-barbell}; see exception \fbadge{7}).
\finding{4}{SemanTok achieves higher decoder-REPA semantic alignment at all noise levels, including the pure noise setting.}
(\cref{fig:dino-probe-main,fig:dino-probe}; see exception \fbadge{7}).
\finding{5}{SemanTok has high fidelity on reconstruction as well as generation.}
\vft{} performs worse on generation than reconstruction, while \ours{} performs well on both (\cref{fig:recon-ar,tab:recon,fig:recon-qual-app,fig:recon-qual-k600}; see exception \fbadge{7}).
\finding{6}{SemanTok's has better generation fidelity from its earlier tokens, which are cheaper to predict.} Under the same AR model, generation at $k{=}16$ costs about 32\% fewer bits per token,
and pixel detail is
deferred to later tokens, not discarded (\cref{fig:two-term,fig:repeats}).
\finding{7}{SemanTok with 1 token per frame struggles to serve every objective.}
At $k{=}1$, \ours{} trails \vft{} in 
semantic alignment
with a mostly clean latent.
Surprisingly, \ours{} still keeps its \classaccadj{} lead at every
budget (\cref{fig:scaling,tab:gen-kinetics,tab:gen-kinetics-sem,tab:gen-uco3d,tab:gen-uco3d-sem,tab:repa}).
\end{findings}

\section{Related Work}
\label{sec:related}

\resultblock{Token-based video generation.}Video models predict discrete tokens from grid tokenizers,
autoregressively~\citep{yan2021videogpt,kondratyuk2024videopoet} or by
masked prediction~\citep{villegas2023phenaki,yu2023language},
and world models use the same recipe for action-conditioned
rollouts~\citep{bruce2024genie,nvidia2025cosmos} or predict
self-supervised features instead of pixels~\citep{vjepa2_2025,dinowm2024}.
In hybrids, AR fixes the content and diffusion renders
it~\citep{li2024mar,li2025arlon}; VAR instead orders prediction
coarse-to-fine across scales~\citep{tian2024var}.
Every grid location still reaches the generator.

\resultblock{Ordered and flexible tokenizers.}TiTok and LARP compress images and videos into 1D token
sequences~\citep{yu2024titok,wang2024larp}.
Nested dropout and Matryoshka losses order such codes by importance~\citep{rippel2014nested,kusupati2022matryoshka}:
ElasticTok drops token suffixes~\citep{yan2024elastictok},
Semanticist finds a PCA-like, semantics-first ordering~\citep{wen2025semanticist},
and FlexTok and VideoFlexTok make every prefix decodable, with REPA on an
early decoder layer~\citep{flextok2025,videoflextok,yu2025repa}.
ReToK strengthens decoder alignment for shorter prefixes~\citep{fu2026retok},
LoST aligns ordered 3D latents with DINO~\citep{dutt2026lost}, and
SpeechTokenizer distills a teacher into its first audio
level~\citep{zhang2024speechtokenizer}.
SemanTok trains every retained prefix of a flexible video code to predict
the teacher without the noised latent, which no prior video tokenizer does to our knowledge.

\resultblock{Tokenizers for generation, not reconstruction.}Reconstruction alone does not make a latent easy to model.
DiGIT, MAETok, VA-VAE, ImageFolder, GigaTok, and UniTok align or predict
foundation-model features to improve
generation~\citep{zhu2024digit,chen2025maetok,yao2025vavae,li2025imagefolder,xiong2025gigatok,ma2025unitok};
REPA-E trains the VAE through the REPA loss~\citep{leng2025repae}, and RAE
diffuses directly in frozen DINO features~\citep{zheng2025rae}.
LARP and CRT shape the code with an AR prior, trading reconstruction for
generation~\citep{wang2024larp,ramanujan2024crt}, consistent with the
perception--distortion trade-off~\citep{blau2018perception}.
These works use fixed-length, mostly image, codes. We place semantics in the \emph{early prefixes} of a flexible video
tokenizer, where a small AR model spends its budget.

\begin{figure}[t]
\centering
\includegraphics[width=\linewidth]{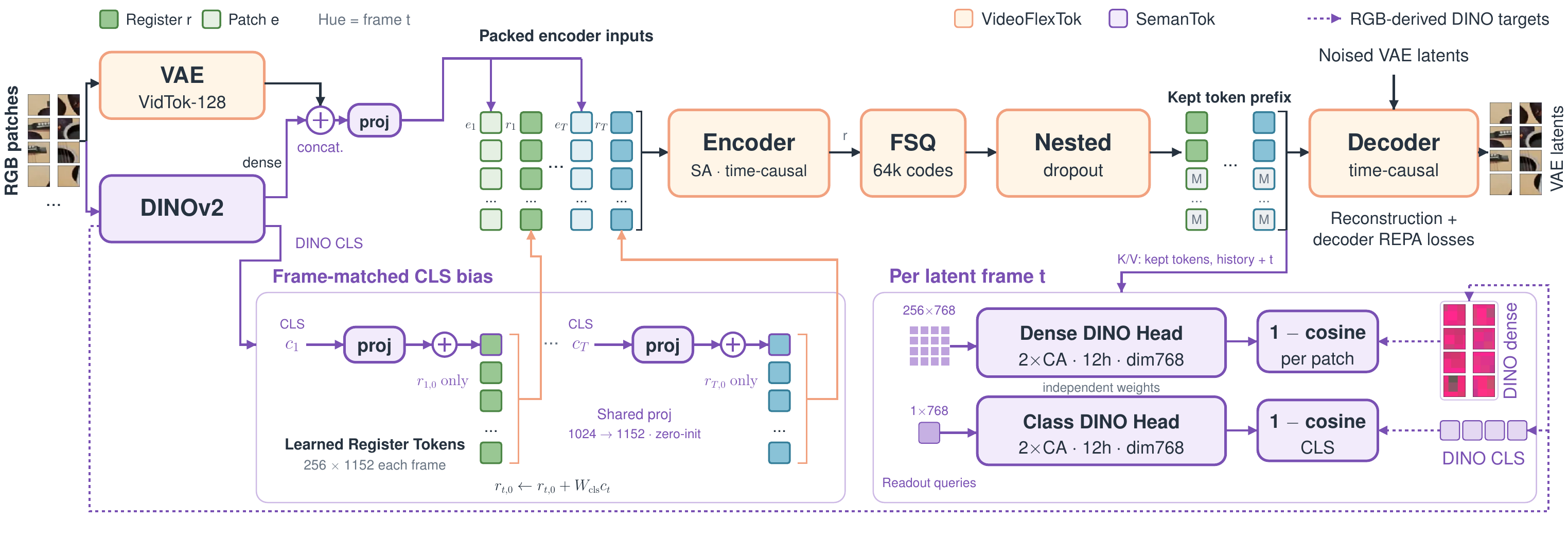}
\vspace{-\parskip}
\vspace{-2.0\abovecaptionskip}
\caption{The figure contrasts two tokenizer variants. \vft{}
provides the base coarse-to-fine path
(\textcolor[HTML]{F1A17F}{orange}), while \ours{} changes
its encoder input and adds a Dense DINO head and a Class DINO head
(\textcolor[HTML]{7950B3}{purple}). A frozen VidTok VAE maps the clip to latents; each frame
projects its patches to $e$ and packs them with $K$ learnable register tokens
$\mathbf r$:
$(e_{t,0},\ldots,e_{t,P-1},\,r_{t,0},\ldots,r_{t,K-1})$, interleaved over
time. The time-causal encoder transforms the packed sequence, retaining
only the register-token outputs for FSQ; nested dropout forms the kept
token prefix that conditions a time-causal decoder reconstructing VAE
latents under flow-matching and decoder-REPA losses. \ours{} concatenates frozen DINOv2 patch features with each
VAE patch before projection, and adds a zero-initialized projection of
the matching DINO class token to $r_{t,0}$. Readout queries in the Dense
DINO head and Class DINO head cross-attend to the kept token prefix from
frames $t'\le t$.}
\label{fig:arch}
\end{figure}

\section{Method}
\label{sec:method}
\abovedisplayskip=3pt
\belowdisplayskip=3pt
\abovedisplayshortskip=2pt
\belowdisplayshortskip=2pt
\setlength{\jot}{1pt}

We review \vft{} (orange in \cref{fig:arch}), then \ours{}.
Our goal is to make the early prefix carry clip semantics without
changing the codebook, native decoder, or AR interface.

\resultblock{VideoFlexTok tokenizer.}
Let $x$ be an RGB clip.
A frozen VidTok VAE maps $x$ to
$z\in\mathbb{R}^{T\times h\times w\times C_z}$, where
$T{=}5$, $h{=}w{=}C_z{=}16$.
We write $\mathbf z_{t,p}\in\mathbb{R}^{C_z}$ for the latent vector at
frame $t$ and spatial position $p$, with
$t\in\{1,\ldots,T\}$ and $p\in\{0,\ldots,P{-}1\}$, where
$P{=}h{\times}w{=}256$.
A learned linear map lifts each patch to encoder width $d_e{=}1152$:
\[
\mathbf e_{t,p}
 =W_{\mathrm{in}}\mathbf z_{t,p}+\mathbf b_{\mathrm{in}}
 \in\mathbb{R}^{d_e}.
\]
Independently, the encoder holds $K{=}256$ learnable register tokens
$\mathbf r_{t,i}\in\mathbb{R}^{d_e}$,
$i\in\{0,\ldots,K{-}1\}$, following \vft{}~\citep{videoflextok}.
Index $i$ is the coarse-to-fine token position.

For each frame the encoder packs the $P$ patches, then the $K$ register tokens:
\[
\mathcal{S}_t
 =(\mathbf e_{t,0},\ldots,\mathbf e_{t,P-1},
  \mathbf r_{t,0},\ldots,\mathbf r_{t,K-1}).
\]
The encoder input is the time-interleaved sequence
$(\mathcal{S}_1,\ldots,\mathcal{S}_T)$.
With time-causal attention, frame $t$ sees only past frames; within each
frame, patches attend freely and register token~$i$ reads all patches and
register tokens $j\le i$.
Patch embeddings are discarded; each register-token output, having already
read the patches, is linearly projected to $\mathbf u_{t,i}\in\mathbb{R}^{D_q}$, $D_q{=}6$, and passed
through FSQ~\citep{mentzer2023fsq}, which bounds the six dimensions with
$\tanh$ and rounds on the lattice $[8,8,8,5,5,5]$ to give
\[
\mathbf q_{t,i}=\operatorname{FSQ}(\mathbf u_{t,i})
\in\mathbb{R}^{D_q}.
\]
Nested dropout samples one $k$ uniformly from $\{1,2,4,\ldots,256\}$ per
clip and replaces $\mathbf q_{t,i}$ for $i\ge k$ in every latent frame with
a learned mask token.
The kept token prefix conditions a time-causal rectified-flow decoder on
noised VAE latents.
\vft{} trains encoder, FSQ, nested dropout, and decoder jointly
(step~(1) in \cref{fig:overview}) with
\[
\mathcal{L}_{\mathrm{base}}
=\mathcal{L}_{\mathrm{Flow}}
+\lambda\,\mathcal{L}^{\mathrm{dec}}_{\mathrm{REPA}},
\qquad \lambda=1,
\]
where $\mathcal{L}_{\mathrm{Flow}}$ is flow matching and the decoder REPA
term aligns an early decoder layer with frozen DINOv2 patches.

Each $\mathbf q_{t,i}$ also maps to a codebook index
\[
\tau_{t,i}=\operatorname{idx}(\mathbf q_{t,i})
\in\{0,\ldots,V{-}1\},
\qquad
V=\prod_j L_j=64000.
\]
A separate AR model (step~(2) in \cref{fig:overview}) is trained on all
$TK$ indices in time-first order,
$(\tau_{1,0},\ldots,\tau_{T,0},\tau_{1,1},\ldots,\tau_{T,K-1})$,
so any prefix is a valid token budget at inference; the frozen tokenizer
decoder renders AR-sampled tokens to video.

\resultblock{SemanTok tokenizer.}
\ours{} keeps the codebook, nested dropout, decoder, and
$\mathcal{L}_{\mathrm{base}}$; the purple paths in \cref{fig:arch} add semantic supervision, which changes the encoder input and supervises the tokens.
A frozen DINOv2-L teacher~\citep{oquab2023dinov2} supplies, per latent
frame, a $16{\times}16$ patch grid $\mathbf d_{t,p}$ aligned with the
$P$ latent positions and a class token $\mathbf c_t$, both in
$\mathbb{R}^{D_D}$, $D_D{=}1024$.
Each patch embedding fuses both inputs,
$\mathbf e_{t,p}=W_{\mathrm{in}}[\mathbf z_{t,p};\mathbf d_{t,p}]
+\mathbf b_{\mathrm{in}}$, and a zero-initialized projection of the class
token is added to the first register token,
$\widetilde{\mathbf r}_{t,0}=\mathbf r_{t,0}+W_{\mathrm{cls}}\mathbf c_t$.
Packing, attention, and FSQ are otherwise as above.

Two independently parameterized cross-attention heads read the kept
token prefix.
For frame $t$, their shared context at budget $k$ is
\[
\mathcal{C}_{t,k}
=\{\mathbf q_{s,i}:1\le s\le t,\ 0\le i<k\}.
\]
The temporal restriction matches the tokenizer's causal path.
Each head has two cross-attention layers of width $d_h{=}768$ with
12 attention heads.
The Dense DINO head $h_\phi$ has spatial readout queries
$\mathbf a_p\in\mathbb{R}^{d_h}$, which each aim to reconstruct one DINO
patch from the linear token sequence, and the Class DINO head
$h_\psi$ has a single readout query $\mathbf a_{\mathrm{cls}}$, distinct
from the register token $\mathbf r_{t,0}$:
\begin{gather*}
\widehat{\mathbf D}_t
=\{\widehat{\mathbf d}_{t,p}\}_{p=0}^{P-1}
=h_\phi(\mathcal{C}_{t,k}), \qquad
\widehat{\mathbf c}_t=h_\psi(\mathcal{C}_{t,k}),\\
\mathcal{L}_{\mathrm{dense}}
=\frac{1}{TP}\sum_{t=1}^{T}\sum_{p=0}^{P-1}
   \left(1-\cos(\widehat{\mathbf d}_{t,p},\mathbf d_{t,p})\right),
\qquad
\mathcal{L}_{\mathrm{cls}}
=\frac{1}{T}\sum_{t=1}^{T}
\left(1-\cos(\widehat{\mathbf c}_t,\mathbf c_t)\right).
\end{gather*}
Both heads supervise the same discrete representation consumed by the
decoder and predicted by the AR model; unlike the decoder REPA
layer, they never see the noised latent, so only the tokens can lower
their losses.
Nested dropout varies $k$, so each sampled prefix must support both DINO
predictions.

The full objective is
\[
\mathcal{L}
=\mathcal{L}_{\mathrm{base}}
+\mu_{\mathrm{dense}}\mathcal{L}_{\mathrm{dense}}
+\mu_{\mathrm{cls}}\mathcal{L}_{\mathrm{cls}},
\qquad
\mu_{\mathrm{dense}}=\mu_{\mathrm{cls}}=0.5.
\]
No loss assigns a particular DINO feature to a particular token.
The ordering emerges from the shared prefix constraint, and the
decoder target remains $\mathbf z$, never DINO features.

\abovedisplayskip=7pt plus 2pt minus 5pt
\belowdisplayskip=\abovedisplayskip
\abovedisplayshortskip=0pt plus 3pt
\belowdisplayshortskip=4pt plus 3pt minus 3pt
\makeatletter
\setlength{\@fpsep}{6pt plus 2pt minus 2pt}
\setlength{\floatsep}{8pt plus 2pt minus 2pt}
\setlength{\textfloatsep}{10pt plus 2pt minus 4pt}
\makeatother
\hyphenation{SemanTok VideoFlexTok}
\section{Experimental Details}
\label{sec:setup}

We compare \ours{} to its closest prior work \vft{} which differs only in semantic supervision, at the same codebook, sequence length, and AR recipe.
We ask how that supervision changes generation.
First, we measure how \fid{fidelity} and \sem{semantic alignment} change with
AR model size~\fbadge{1}, how larger and longer-trained AR models improve
fidelity further~\fbadge{2}, and whether semantic alignment holds on
out-of-distribution classes~\fbadge{3}.
We then examine how the token prefix affects decoder-REPA semantic
alignment~\fbadge{4}.
We measure the difference between generation fidelity and reconstruction fidelity i.e. realization gap~\fbadge{5}, and trace \ours{}'s generation
gain to specific tokens~\fbadge{6}.
Finally, we report where \ours{} falls behind: at one token per
frame~\fbadge{7}.



\resultblock{Data}We use Kinetics-600 for
class-to-video setting with a controlled action label, and uCO3D for text-to-video over objects including an
in-distribution (ID) and out-of-distribution (OOD) (never seen in training) class split.

\resultblock{Model}
Following \vft{}~\citep{videoflextok}, \ours{} shares the
backbone, decoder REPA, codebook, sequence length, sampling settings, LLaMA-style AR recipe, and pixel resolution
(17 frames at $128{\times}128$ on both datasets).
\ours{} changes the encoder input and adds the token-level objectives
of \cref{sec:method} using
DINOv2.

\resultblock{Evaluation}
The metrics we evaluate SemanTok on are:
\begin{itemize}[leftmargin=1.2em,itemsep=1pt,topsep=2pt,parsep=0pt]
\item \fid{\textbf{Fidelity}} i.e. closeness in appearance to real video,
measured by \gFVD{} and \gFID{}.
\item \sem{\textbf{Semantic alignment}} i.e. whether the video shows the
conditioned content: \classacc{}, text--video cosine similarity (\ViCLIP{}), video--video cosine similarity in \ViCLIP{} space (\ClipV{}). For the tokenizer we also report \decalign{} (REPA), the
DINOv2 cosine similarity of the decoder-REPA readout.
\end{itemize}
On Kinetics-600, \classacc{} is closed-set UMT-L
top-1~\citep{li2023umt} over 2{,}048 generated clips.
On uCO3D, it is nearest-class-mean top-1 in InceptionV3 space over
2{,}560 clips per split, which stays defined for held-out classes.
Additional details, including bootstrap intervals,
the scope of our ablations, and the tokenizer training budget, are in
\cref{sec:app-hparams}.

\resultblock{Scalability and efficiency.}
We sweep seven AR sizes from 49M to 2.29B parameters, their inference
FLOPs, and a 1.33B AR model trained from 1.3B to 65.5B tokens.
We measure efficiency at two levels.
At the model level, it is the AR size needed to reach a given
\fid{fidelity} and \sem{semantic alignment}.
At the token level, it is how hard the tokens are to predict: the validation
cross-entropy (in bits per token).
All AR models train on all 256 tokens per latent frame, and the budget
$k\in\{1,2,4,\ldots,256\}$ is chosen only at evaluation.

\section{Results}
\label{sec:results}

\begin{figure}[t]
\centering
\includegraphics[width=\textwidth]{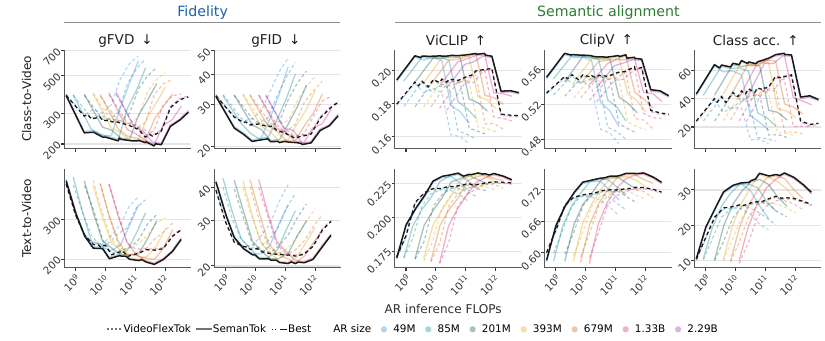}
\vspace{-\parskip}
\vspace{-2.0\abovecaptionskip}
\caption{\fbadge{1}~\ours{} improves generation at every compute budget on Kinetics-600 and uCO3D.
Generation versus AR inference FLOPs per clip.
Each faded curve is one AR size (colour) sweeping the token budget $k$
from 1 to 256 tokens per frame.
Black: the best score each tokenizer reaches at a given compute.
\ours{}'s envelope is better over most of the compute range.}
\label{fig:scaling}
\end{figure}

\begin{figure}[t]
\begin{minipage}[t]{0.47\linewidth}
\vspace{0pt}
\centering
\includegraphics[width=\linewidth]{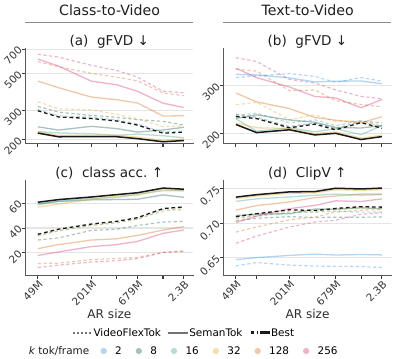}
\vspace{-\parskip}
\vspace{-2.0\abovecaptionskip}
\caption{\small \fbadge{2} Fidelity scales with AR model size for both VideoFlexTok and SemanTok, while the semantic-alignment gap between does not close.}
\label{fig:ar-size}
\end{minipage}\hfill
\begin{minipage}[t]{0.492\linewidth}
\vspace{0pt}
\centering
\includegraphics[width=\linewidth]{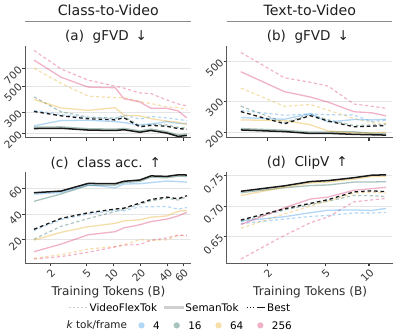}
\vspace{-\parskip}
\vspace{-2.0\abovecaptionskip}
\caption{\small \fbadge{2} Semantic alignment gap between SemanTok and VideoFlexTok closes slower than the fidelity gap between them.}
\label{fig:trainD}
\end{minipage}
\end{figure}

\findinghead{1}{SemanTok exhibits high semantic alignment and video fidelity at every AR model size.}With each tokenizer at its best $k$, every \ours{} AR model reaches lower
\gFVD{} and \gFID{} and higher \classacc{} than the \vft{} AR model of the
same size, on both Kinetics-600 and uCO3D (\cref{fig:scaling,fig:ar-size}).
On Kinetics-600, \ours{} lowers \gFVD{} by 11--24\% and \gFID{} by 5--17\%
and raises \classacc{} by 25--61\%, with the largest gains for the smallest
AR models.
On uCO3D, \ours{} improves \gFVD{}, \gFID{}, and \classacc{} by 2--13\%,
5--8\%, and 22--30\%, with no clear trend in AR model size.
At the single budget $k{=}16$, an 85M \ours{} AR model beats
\vft{} AR model of every size and budget, up to 2.29B, in \gFID{} on both
datasets and in \gFVD{} on Kinetics-600.
The 85M \ours{} AR model matches that \vft{} AR model in uCO3D \gFVD{} and
beats it in \classacc{} on both datasets.

In \sem{semantic alignment} even the smallest \ours{} model (49M) beats the best \vft{} AR model of any size, up to 2.29B, in class accuracy, ClipV, and ViCLIP on both datasets. On Kinetics-600, \ours{}'s class accuracy is 0.631 against 0.560 for the 2.29B \vft{} model.

More tokens cost more AR
compute and their generations degrade in gFVD beyond a certain point. However, \vft{} gives worse \gFVD{} after only $k{=}16$, while \ours{} does not degrade until after $k{=}32$.

\begin{figure*}[t]
\centering
\includegraphics[width=\textwidth]{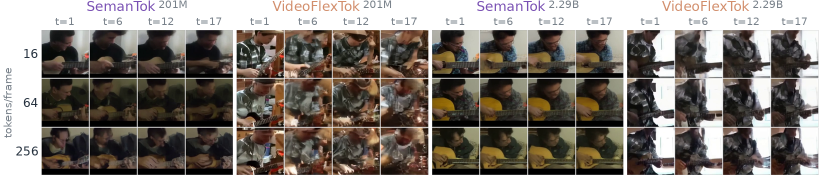}
\vspace{-\parskip}
\vspace{-2.0\abovecaptionskip}
\caption{\fbadge{2}~\ours{} stays ahead at both AR model sizes.Class-to-video rollouts for one Kinetics-600
playing-guitar label.
At 201M, \ours{} degrades slowly over time, while \vft{} is sharp
only at $t{=}1$.
Videos on the project page.}
\label{fig:guitar-seated}
\end{figure*}

\findinghead{2}{Increasing SemanTok model size further improves fidelity.}Larger or longer-trained AR models mostly trim the compounding error of the
token rollout, which improves \fid{fidelity}.
The same scaling gains less in \sem{semantic alignment}, because what the clip shows is
largely decided by the first tokens, which come from the tokenizer
(\fbadge{6}).
\Cref{fig:ar-size} separates the two axes by AR size.
\ours{}'s best-$k$ \classacc{} and \ClipV{} lie above
\vft{} at every size.
\fid{Fidelity} improves with AR model size for both tokenizers, and \ours{}
reaches a given \fid{fidelity} with fewer parameters.
From 49M to 2.29B, \ours{}'s best-$k$ \gFVD{} falls from $224$ to $202$ on
Kinetics-600 and from $218$ to $197$ on uCO3D.
\ours{}'s benefit is largest where AR capacity is scarce: at $k{=}64$,
\ours{}'s \gFID{} gain over \vft{} halves from 49M to 2.29B.
Training budget behaves like model size (\cref{fig:trainD}).
For the 1.33B AR model on Kinetics-600, \ours{}'s best-$k$ \gFVD{} lead shrinks from $28\%$ to $9\%$ by $26$B
tokens, then holds at $11$--$14\%$.
\ours{}'s \classaccadj{} lead on Kinetics-600 persists, at $30\%$ after
$65.5$B tokens.
So $5\times$ more training buys \vft{} some \fid{fidelity}, but not \sem{semantic alignment}.
On uCO3D, \ours{}'s \ClipV{} nearly saturates by $13$B tokens and stays
above \vft{}'s.
As one rollout demonstrates (\cref{fig:guitar-seated}), scaling to 2.29B sharpens the videos of both tokenizers, but \ours{}'s maintains the lead.

\begin{figure}[t]
\centering
\includegraphics[width=0.9\textwidth]{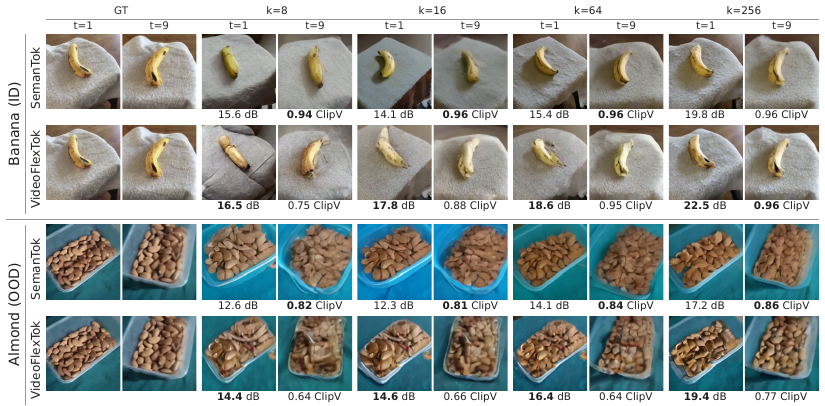}
\vspace{-\parskip}
\vspace{-1.0\abovecaptionskip}
\caption{\fbadge{3}~\ours{} reconstructions on uCO3D from
ground-truth tokens recover the original object class at an earlier $k$ than VideoFlexTok; for both in-distribution (ID) clips and 
out-of-distribution (OOD) clips.}
\label{fig:recon-idood}
\end{figure}

\begin{figure}[t]
\begin{minipage}[t]{0.492\linewidth}
\vspace{0pt}
\centering
\includegraphics[width=\linewidth]{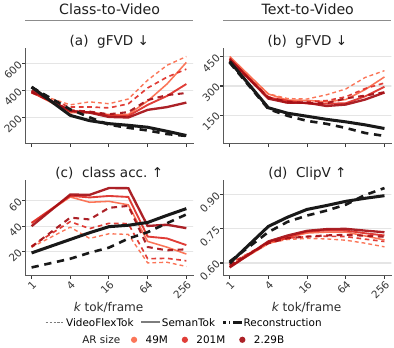}
\vspace{-\parskip}
\vspace{-2.0\abovecaptionskip}
\caption{\small \fbadge{5}~ \ours{} loses less fidelity and semantic alignment from reconstruction to generation than \vft{}.}
\label{fig:recon-ar}
\end{minipage}\hfill
\begin{minipage}[t]{0.492\linewidth}
\vspace{0pt}
\centering
\footnotesize
\setlength{\tabcolsep}{1.6pt}
\setlength{\aboverulesep}{2pt}
\setlength{\belowrulesep}{2pt}
\newlength{\cataccblockht}
\setlength{\cataccblockht}{2.32em}
\newcommand{\cataccmetric}[1]{%
  \multirow{2}{*}{%
    \begin{minipage}[c][\cataccblockht][c]{0.125in}%
      \centering\rotatebox{90}{\scriptsize #1}%
    \end{minipage}}}
\begin{tabular}{@{}>{\centering\arraybackslash}m{0.125in}@{}lccc@{}}
\multicolumn{2}{@{}l}{} & $k{=}4$ & $k{=}16$ & $k{=}256$ \\
\midrule
\cataccmetric{\sem{Acc.}} & \rule{0pt}{1.06em}\vft{} & $0.23$/$0.16$ & $0.27$/$0.18$ & $0.24$/$0.16$ \\
 & \rule{0pt}{1.06em}\ours{} & $\mathbf{0.27}$/$\mathbf{0.18}$ & $\mathbf{0.34}$/$\mathbf{0.23}$ & $\mathbf{0.27}$/$\mathbf{0.18}$ \\
\midrule
\cataccmetric{\sem{ViCLIP}} & \rule{0pt}{1.06em}\vft{} & $\mathbf{0.21}$/$\mathbf{0.19}$ & $0.23$/$0.20$ & $0.22$/$0.19$ \\
 & \rule{0pt}{1.06em}\ours{} & $0.21$/$0.18$ & $\mathbf{0.23}$/$\mathbf{0.20}$ & $\mathbf{0.23}$/$\mathbf{0.20}$ \\
\midrule
\cataccmetric{\sem{ClipV}} & \rule{0pt}{1.06em}\vft{} & $0.70$/$\mathbf{0.63}$ & $0.73$/$0.65$ & $0.70$/$0.63$ \\
 & \rule{0pt}{1.06em}\ours{} & $\mathbf{0.70}$/$0.62$ & $\mathbf{0.75}$/$\mathbf{0.66}$ & $\mathbf{0.73}$/$\mathbf{0.65}$ \\
\end{tabular}
\vspace{-\parskip}
\vspace{2.5pt}
\captionof{table}{\small \fbadge{3}The semantic gain holds on
unseen classes. \sem{Semantic alignment} of generated videos from a 201M AR model for in-distribution (ID) and out-of-distribution (OOD) samples of uCO3D dataset; bold is better.
\ours{} leads \sem{class accuracy} on both splits at every $k$.}
\label{tab:catacc}
\end{minipage}
\end{figure}

\findinghead{3}{SemanTok is able to maintain semantic alignment over out-of-distribution classes.}We test this on uCO3D, whose out-of-distribution (OOD) object classes are
never seen in training, unlike its in-distribution (ID) classes.
\ours{} leads on \sem{semantic-alignment} in both tokenizer reconstruction and AR generation.
\ours{} prefixes recover the object class at lower $k$ better than \vft{}'s
on both ID and OOD clips (\cref{fig:recon-idood}).
At $k{=}16$ on uCO3D, \ours{}'s reconstructions score higher \ClipV{} and
\classacc{} than \vft{}'s but about $2.7$\,dB lower \PSNR{}
(\cref{tab:recon}).
The lead carries over to AR generation:
with a 201M AR model at $k{=}16$, \ours{} raises generated \classacc{} over
\vft{} by $24\%$ on ID clips and by $29\%$ on OOD classes, and improves
\ClipV{} on both
(\cref{tab:catacc}).
\ViCLIP{} favors \ours{} from $k{=}16$ on, but not at $k{=}4$.

\begin{figure}[t]
\centering
\includegraphics[width=\linewidth]{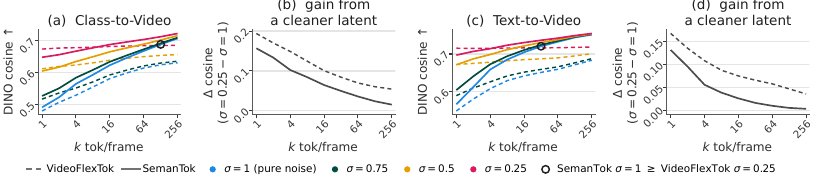}
\vspace{-\parskip}
\vspace{-2.0\abovecaptionskip}
\caption{\small \fbadge{4}~\ours{}'s decoder-REPA readout depends less on the
noised latent and achieves higher REPA-alignment than \vft{}'s.}
\label{fig:dino-probe-main}
\end{figure}

\findinghead{4}{SemanTok achieves higher decoder-REPA semantic alignment at all noise levels, including the pure noise setting.}Decoder REPA aligns an early decoder layer with DINOv2 features(\cref{sec:method}).
The decoder also sees a partly noised latent, which can supply part of this
target without the tokens.
We therefore read out the REPA projection on both
datasets while varying the noise level $\sigma$ of the decoder input
(\cref{fig:dino-probe-main}).
For generation from pure noise, \ours{}'s readout has a higher DINOv2 cosine similarity than \vft{}'s at every $k$.
At $k{=}32$ on uCO3D, \ours{}'s pure-noise readout reaches the similarity that \vft{}
reaches only with a $75\%$-clean latent ($0.721$ vs.\ $0.716$).
At $k{=}256$ the latent adds almost nothing to \ours{}'s readout, but
over $11\times$ more to \vft{}'s (\cref{sec:app-dino-probe}). (The exception is the smallest budgets with a mostly clean latent, see \fbadge{7}).

\findinghead{5}{SemanTok has high fidelity on reconstruction as well as generation} (see \cref{fig:recon-ar}).
\vft{} reconstructs somewhat better in \PSNR{} and \rFVD{}, but it loses more
\fid{fidelity} in generation, so its realization gap is wider.
\ours{} has the lower \gFVD{} at every AR size, at both $k{=}64$ and
$k{=}128$, and hence lower realization gap.
(On Kinetics-600, generated \classacc{} can exceed reconstruction because the AR model sees the class label.)

\begin{figure*}[ht]
\centering
\includegraphics[width=\textwidth]{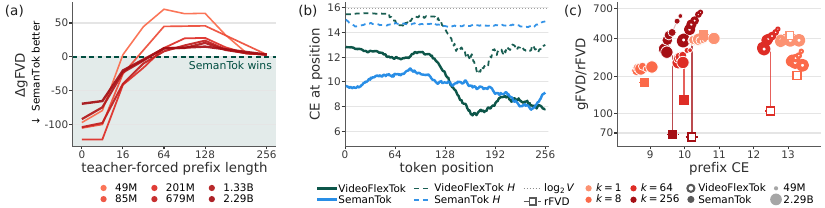}
\vspace{-\parskip}
\vspace{-2.0\abovecaptionskip}
\caption{\fbadge{6}~\ours{}'s generation gain comes specifically from its first tokens, which are cheaper to predict (Kinetics-600).
(a)~$\Delta$\gFVD{} after teacher-forcing the first $m$ ground-truth tokens
per frame and free-running to $k{=}256$: at 201M, \ours{} leads by $23\%$
with nothing forced, and forcing only the first 16--64 tokens removes the
lead.
(b)~Under a 201M AR model, \ours{}'s cross-entropy per token position is
lower for the first 128 positions and higher over 129--192.
(c)~Larger AR models narrow the gap between generation and reconstruction
for both tokenizers, but none closes it.}
\label{fig:two-term}
\end{figure*}

\findinghead{6}{SemanTok's generation fidelity gain comes from its earlier tokens, which are cheaper to predict.}Two
measurements on Kinetics-600 show that \ours{}'s gain is concentrated in the earlier tokens.
First, we force the first $m$ ground-truth tokens per frame and free-run to
$k{=}256$ (\cref{fig:two-term}\captiona).
With nothing forced, \ours{}'s \gFVD{} is $23\%$ lower than \vft{}'s at
201M.
\ours{}'s lead vanishes once the first 16--64 tokens are forced, after which
\vft{}'s better-reconstructing tail edges ahead.
Second, \ours{}'s prefix is cheaper to predict (\cref{fig:two-term}\captionb).

The bits the AR model needs per predicted token is measured by cross-entropy.
At $k{=}16$, \ours{} needs $32\%$ fewer bits per token than \vft{} ($8.8$
vs.\ $12.9$ at 201M).
\ours{}'s marginal entropy is only about one bit lower, so most of the saving comes from context.
It is to be noted that the cost saving does not come from repetition: \ours{} repeats tokens less often
overall than \vft{} (\cref{sec:app-repeats}).

\ours{}'s cheap prefix keeps the semantics and defers pixel detail.
\ours{}'s first 4 tokens cost $37$ bits per frame, yet nearly match the
\classacc{} of \vft{}'s first 32 tokens ($405$ bits,
\cref{tab:recon}).
Later tokens restore part of that detail, and the generative decoder fills
in the rest.

\findinghead{7}{SemanTok with one token per frame struggles to serve every objective.}This fact is consistent throughout our experiments
(\cref{tab:gen-kinetics,tab:gen-kinetics-sem,tab:gen-uco3d,tab:gen-uco3d-sem,tab:repa}).
\ours{}'s losses to \vft{} that clear the bootstrap intervals are all on uCO3D at $k{\le}4$, mostly \gFID{} and \ClipV{} at $k{=}1$.
On Kinetics-600, \ours{} is never significantly worse.
Decoder REPA shows the same budget limit (\cref{fig:dino-probe-main}): at
$k{=}1$, \ours{}'s pure-noise readout leads by only $0.01$--$0.02$ DINOv2 cosine,
against $0.04$--$0.08$ from $k{=}16$, and with a mostly clean latent
($\sigma{=}0.25$) \ours{}'s readout is lower up to $k{=}4$ on uCO3D and
$k{=}8$ on Kinetics-600.
A token carries at most about $16$ bits, too few to satisfy flow matching,
decoder REPA, and semantic supervision at the same time.
\ours{} still never falls behind in \classacc{}, at any budget or AR size,
and \ours{}'s \fid{fidelity} lead follows at larger budgets \fbadge{1}.


\textbf{Limitation} : Because SemanTok prioritizes semantic alignment over reconstruction, it struggles to reconstruct the same colors/appearance details at lower token budgets.

\section{Conclusion}
\label{sec:conclusion}

We introduce \ours{}, an AR video generation tokenizer and prediction module, which emphasizes semantic alignment at flexible token budget.
Semantic supervision from a frozen teacher, as encoder input and as a target for every retained prefix, makes token prefixes richer in semantics (\fbadge{1}), lends to scaling in size (\fbadge{2}), maintains generalization in out-of-distribution setting (\fbadge{3}), has better semantic alignment in high noise inputs (\fbadge{4}), performs well at both reconstruction and generation (\fbadge{5}), and is cheaper to predict (\fbadge{6}).

Future directions include optimizing predictability directly, organizing
early prefixes with video-native or language-aligned teachers, and using
such prefixes as compact states for action-conditioned world models.


\section*{Reproducibility Statement}
Both tokenizers share the architecture and objective in \cref{sec:method}.
SemanTok's changes are the DINO encoder input, the class-token injection,
and the two prediction heads with their loss weights.
\Cref{tab:tok-hparams} lists tokenizer settings and training budgets.
\Cref{tab:ar-size-hparams,tab:ar-data-hparams} list the AR size ladder,
learning rates, and dataset-specific regularization.
\Cref{sec:app-hparams} gives the sampling settings, including the
budget-dependent AR guidance.
It also describes each metric and its readout model: FVD, FID, \ViCLIP{},
\ClipV{}, UMT-L, and the nearest-class-mean protocol for held-out classes.
All datasets and pretrained models are public and cited where they are
first used.

\bibliographystyle{semantok}
\bibliography{references}

\appendix
\section{Qualitative examples}
\label{sec:app-qual}

Stills from reconstructions (ground-truth tokens through each frozen decoder)
and from AR-model rollouts. Every $k$ column is a prefix of the same token
sequence: the clip's encoder tokens for reconstruction, one rollout's tokens
for generation. Stills cannot show temporal consistency, so we recommend
watching the videos on the project page (\projecturl): each caption names its
video file, and the page holds further examples.

\begin{figure}[H]
\centering
\includegraphics[width=\textwidth]{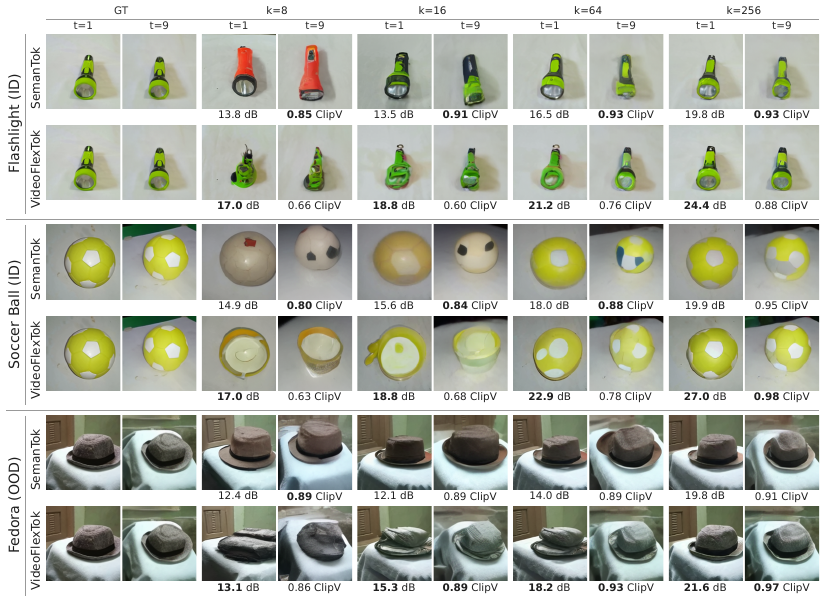}
\caption{Tokenizer reconstruction on uCO3D from ground-truth tokens: two
in-distribution clips and one out-of-distribution clip. PSNR and ClipV against
VAE-GT; bold is best per $k$. Videos: \texttt{ReconstructionUCO3D.mp4},
\texttt{ReconstructionUCO3D\_Flashlight.mp4}.}
\label{fig:recon-qual-app}
\end{figure}

\begin{figure}[H]
\centering
\includegraphics[width=\textwidth]{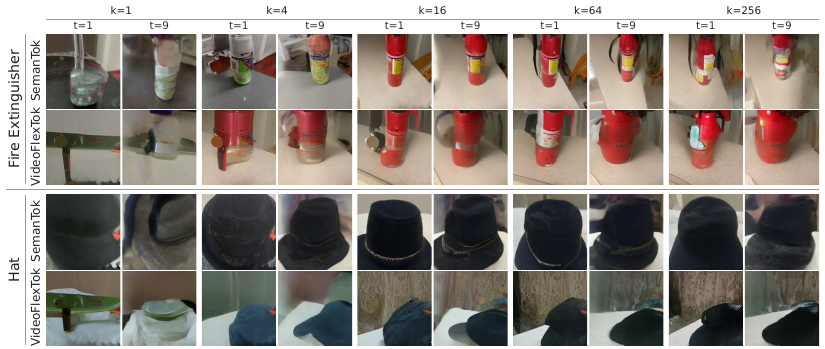}
\caption{Text-to-video on uCO3D, 201M AR model; one rollout per tokenizer,
decoded from its first $k$ tokens per frame. Prompts: ``A red fire
extinguisher with a black handle and silver nozzle, sits on a countertop. It
has a white label on its side and a black strap around its middle.'' and ``A
black fedora hat with a short brim and small, round crown sits on a white
surface. It has two small holes at the top of the crown.''
Video: \texttt{GenerationUCO3D\_FireExtinguisher.mp4}.}
\label{fig:gen-qual-uco3d}
\end{figure}

\begin{figure}[H]
\centering
\includegraphics[width=\textwidth]{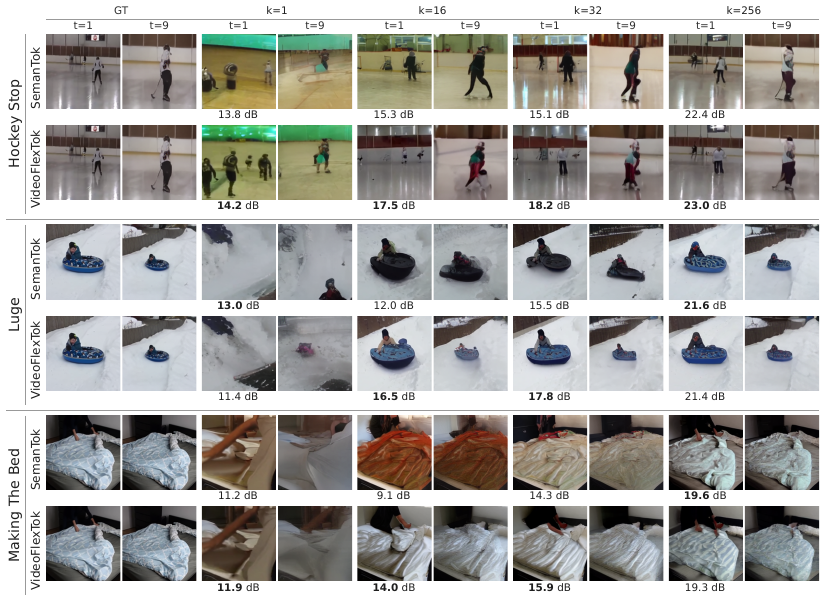}
\caption{Tokenizer reconstruction on Kinetics-600 from ground-truth tokens.
PSNR against VAE-GT; bold is best per $k$. Videos:
\texttt{ReconstructionK600\_HockeyStop.mp4}, \texttt{ReconstructionK600\_Luge.mp4}.}
\label{fig:recon-qual-k600}
\end{figure}

\begin{figure}[H]
\centering
\includegraphics[width=\textwidth]{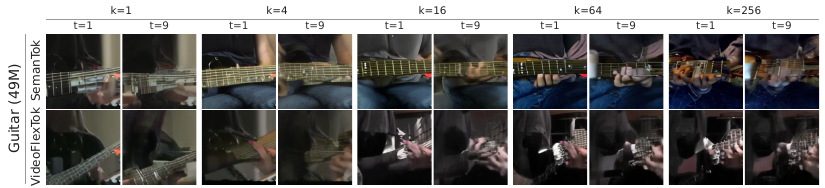}
\caption{Class-to-video on Kinetics-600 (playing guitar) with a 49M AR model;
one rollout per tokenizer (the first evaluation sample of the class), decoded
from its first $k$ tokens per frame. Video:
\texttt{GenerationK600\_Guitar\_49M.mp4}.}
\label{fig:gen-qual-k600}
\end{figure}

\begin{figure}[H]
\centering
\includegraphics[width=\textwidth]{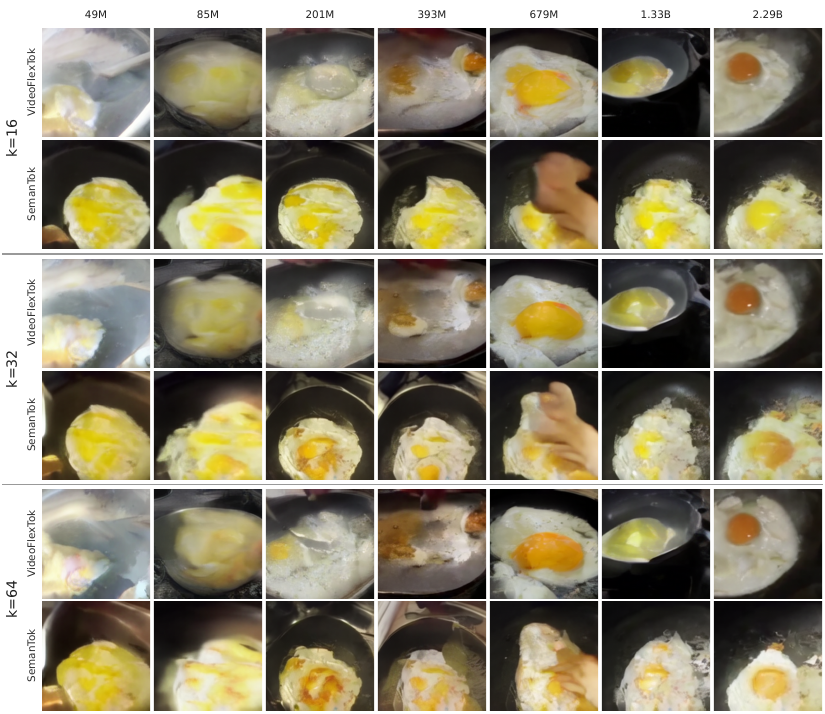}
\caption{Class-to-video on Kinetics-600 (cooking egg) across AR model sizes at
$k{=}16,32,64$, frame $t{=}9$. Each (size, tokenizer) is one rollout decoded
from its first $k$ tokens per frame. SemanTok's is the first evaluation sample
of the class; VideoFlexTok's is, per size, the one of its rollouts of the class
closest to SemanTok's in ViCLIP embedding. Video:
\texttt{GenerationK600\_CookingEgg\_k16-64\_ARsize.mp4}.}
\label{fig:gen-qual-egg}
\end{figure}

\begin{figure}[H]
\centering
\includegraphics[width=\textwidth]{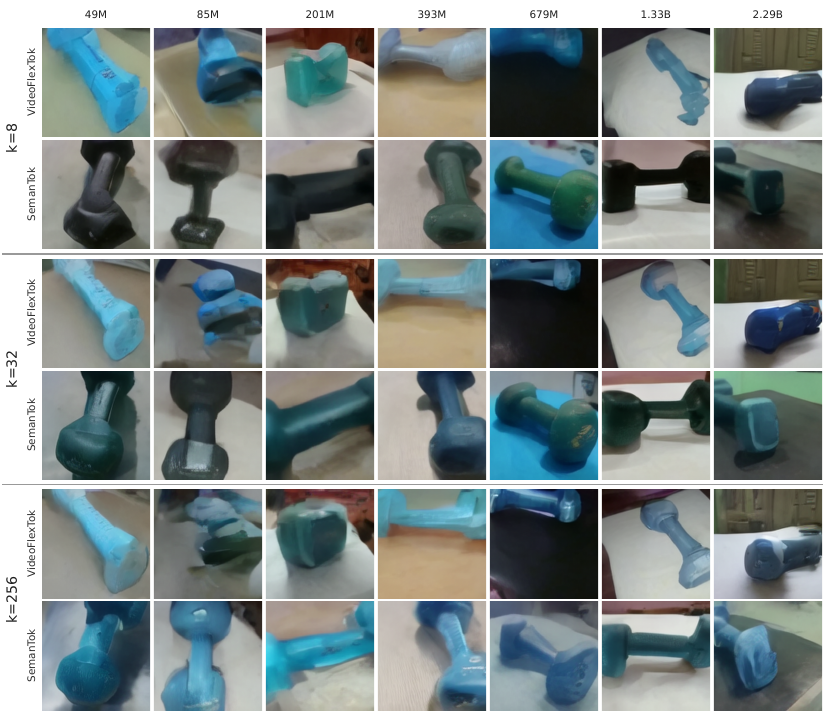}
\caption{Text-to-video on uCO3D (out-of-distribution category) across AR model
sizes at $k{=}8,32,256$, frame $t{=}9$. Each (size, tokenizer) is that model's
evaluation rollout for the same prompt, decoded from its first $k$ tokens per
frame. Prompt: ``A light blue dumbbell with a hexagonal handle and a rounded
head sits on a surface. It has white text or numbers printed on its side, such
as `Tone' or `5 LB'. The overall color scheme is dominated by the blue of the
dumbbell itself.'' Video: \texttt{GenerationUCO3D\_Barbell\_ARsize\_x\_k.mp4}.}
\label{fig:gen-qual-barbell}
\end{figure}

\section{Tokenizer reconstruction versus token budget}
\label{sec:app-recon}
All rows score the same fake side --- the frozen decoder applied to
ground-truth tokens. These are tokenizer reconstruction measurements, not
the AR-model rollouts in \cref{fig:scaling}. Only the reference differs, as
each metric requires: rFVD against the reference distribution, ViCLIP
against the caption, and PSNR/SSIM per clip against VAE- GT. Compare within
a row; the metrics are not comparable across references.
\Cref{fig:recon-ar} overlays the AR-model curves from
\cref{fig:scaling} on these reconstruction numbers.

\begin{table}[h]
\caption{Tokenizer reconstruction versus $k$. Bold is better.}
\label{tab:recon}
\centering
\footnotesize
\setlength{\tabcolsep}{2.7pt}
\begin{tabular}{@{}ll *{8}{c}@{}}
\toprule
&& $1$ & $4$ & $8$ & $16$ & $32$ & $64$ & $128$ & $256$ \\
\midrule
\multicolumn{10}{@{}l}{\textbf{Kinetics-600}} \\
$rFVD\downarrow$ & \vft{}
  & $\mathbf{426.7}$ & $262.0$ & $203.0$ & $\mathbf{148.1}$
  & $\mathbf{125.2}$ & $\mathbf{105.5}$ & $\mathbf{79.1}$ & $\mathbf{64.8}$ \\
& \ours{}
  & $427.8$ & $\mathbf{217.4}$ & $\mathbf{178.3}$ & $156.0$
  & $143.9$ & $129.3$ & $99.6$ & $68.2$ \\[\smallskipamount]
ViCLIP$\uparrow$ & \vft{}
  & $0.155$ & $0.163$ & $0.168$ & $0.174$
  & $0.177$ & $0.182$ & $0.185$ & $0.188$ \\
& \ours{}
  & $\mathbf{0.170}$ & $\mathbf{0.179}$ & $\mathbf{0.182}$ & $\mathbf{0.185}$
  & $\mathbf{0.186}$ & $\mathbf{0.187}$ & $\mathbf{0.190}$ & $\mathbf{0.192}$ \\[\smallskipamount]
ClipV$\uparrow$ & \vft{}
  & $0.569$ & $0.631$ & $0.678$ & $0.730$
  & $0.767$ & $\mathbf{0.804}$ & $\mathbf{0.840}$ & $0.867$ \\
& \ours{}
  & $\mathbf{0.601}$ & $\mathbf{0.681}$ & $\mathbf{0.715}$ & $\mathbf{0.748}$
  & $\mathbf{0.775}$ & $0.803$ & $0.839$ & $\mathbf{0.873}$ \\[\smallskipamount]
class acc.$\uparrow$ & \vft{}
  & $0.074$ & $0.143$ & $0.190$ & $0.230$
  & $0.302$ & $0.353$ & $0.423$ & $0.493$ \\
& \ours{}
  & $\mathbf{0.188}$ & $\mathbf{0.295}$ & $\mathbf{0.348}$ & $\mathbf{0.396}$
  & $\mathbf{0.407}$ & $\mathbf{0.429}$ & $\mathbf{0.485}$ & $\mathbf{0.539}$ \\[\smallskipamount]
PSNR$\uparrow$ & \vft{}
  & $\mathbf{10.24}$ & $\mathbf{11.55}$ & $\mathbf{13.17}$ & $\mathbf{14.64}$
  & $\mathbf{15.80}$ & $\mathbf{17.15}$ & $\mathbf{18.53}$ & $\mathbf{19.66}$ \\
& \ours{}
  & $10.13$ & $10.51$ & $10.78$ & $11.33$
  & $13.02$ & $14.78$ & $16.79$ & $18.53$ \\[\smallskipamount]
SSIM$\uparrow$ & \vft{}
  & $\mathbf{0.278}$ & $\mathbf{0.319}$ & $\mathbf{0.370}$ & $\mathbf{0.417}$
  & $\mathbf{0.466}$ & $\mathbf{0.529}$ & $\mathbf{0.607}$ & $\mathbf{0.668}$ \\
& \ours{}
  & $0.277$ & $0.288$ & $0.297$ & $0.316$
  & $0.372$ & $0.446$ & $0.543$ & $0.634$ \\
\midrule
\multicolumn{10}{@{}l}{\textbf{uCO3D}} \\
$rFVD\downarrow$ & \vft{}
  & $\mathbf{420.8}$ & $\mathbf{181.1}$ & $\mathbf{146.2}$ & $\mathbf{120.1}$
  & $\mathbf{105.1}$ & $\mathbf{83.7}$ & $\mathbf{57.6}$ & $\mathbf{42.4}$ \\
& \ours{}
  & $427.4$ & $187.7$ & $159.6$ & $144.0$
  & $128.2$ & $115.2$ & $99.7$ & $80.8$ \\[\smallskipamount]
ViCLIP$\uparrow$ & \vft{}
  & $0.157$ & $0.194$ & $0.206$ & $0.213$
  & $0.218$ & $0.224$ & $0.234$ & $\mathbf{0.241}$ \\
& \ours{}
  & $\mathbf{0.166}$ & $\mathbf{0.206}$ & $\mathbf{0.214}$ & $\mathbf{0.227}$
  & $\mathbf{0.231}$ & $\mathbf{0.237}$ & $\mathbf{0.237}$ & $0.239$ \\[\smallskipamount]
ClipV$\uparrow$ & \vft{}
  & $\mathbf{0.606}$ & $0.736$ & $0.781$ & $0.808$
  & $0.830$ & $0.856$ & $\mathbf{0.898}$ & $\mathbf{0.929}$ \\
& \ours{}
  & $0.599$ & $\mathbf{0.761}$ & $\mathbf{0.801}$ & $\mathbf{0.835}$
  & $\mathbf{0.852}$ & $\mathbf{0.871}$ & $0.884$ & $0.895$ \\[\smallskipamount]
class acc.$\uparrow$ & \vft{}
  & $0.092$ & $0.315$ & $0.451$ & $0.520$
  & $0.582$ & $0.648$ & $\mathbf{0.733}$ & $\mathbf{0.780}$ \\
& \ours{}
  & $\mathbf{0.111}$ & $\mathbf{0.378}$ & $\mathbf{0.492}$ & $\mathbf{0.565}$
  & $\mathbf{0.619}$ & $\mathbf{0.659}$ & $0.694$ & $0.719$ \\[\smallskipamount]
PSNR$\uparrow$ & \vft{}
  & $\mathbf{13.46}$ & $\mathbf{16.09}$ & $\mathbf{17.34}$ & $\mathbf{18.66}$
  & $\mathbf{19.73}$ & $\mathbf{20.82}$ & $\mathbf{22.59}$ & $\mathbf{23.88}$ \\
& \ours{}
  & $12.49$ & $14.28$ & $15.09$ & $15.96$
  & $17.39$ & $18.86$ & $20.49$ & $21.46$ \\[\smallskipamount]
SSIM$\uparrow$ & \vft{}
  & $\mathbf{0.491}$ & $\mathbf{0.535}$ & $\mathbf{0.565}$ & $\mathbf{0.596}$
  & $\mathbf{0.629}$ & $\mathbf{0.672}$ & $\mathbf{0.734}$ & $\mathbf{0.780}$ \\
& \ours{}
  & $0.464$ & $0.509$ & $0.525$ & $0.544$
  & $0.577$ & $0.619$ & $0.676$ & $0.717$ \\
\bottomrule
\end{tabular}
\end{table}

\section{Where the tokenizers keep DINO semantics}
\label{sec:app-dino-probe}
Both tokenizers train an early decoder layer to predict DINOv2 patch
features through decoder REPA.
We read that projection out without any training, which asks where the
decoder finds its semantics.
We keep the first $k$ tokens per frame and fix the decoder noise level
$\sigma$, with the same noise seed for both tokenizers and every $k$.
At $\sigma{=}1$ the decoder input is pure noise, so any DINO content in the
readout comes from the token prefix.
Generation starts from this state, with AR-sampled tokens as the only
input.
At lower $\sigma$ the noised latent also carries the clip; we measure
$\sigma\in\{1,0.75,0.5,0.25\}$.
We report the mean cosine to the true DINOv2 features of each latent
frame's first RGB frame, the target both tokenizers train against.
Each dataset uses 256 validation clips and the tokenizers of
\cref{sec:results} (66B training tokens on uCO3D, 131B on Kinetics-600), with
SemanTok's class-token input fed as in training.
As a check, at $k{=}256$ and $\sigma{=}0.25$ the readout lies within
$0.05$ of the alignment each trainer logged for the same checkpoint.

\begin{figure*}[ht]
\centering
\includegraphics[width=\textwidth]{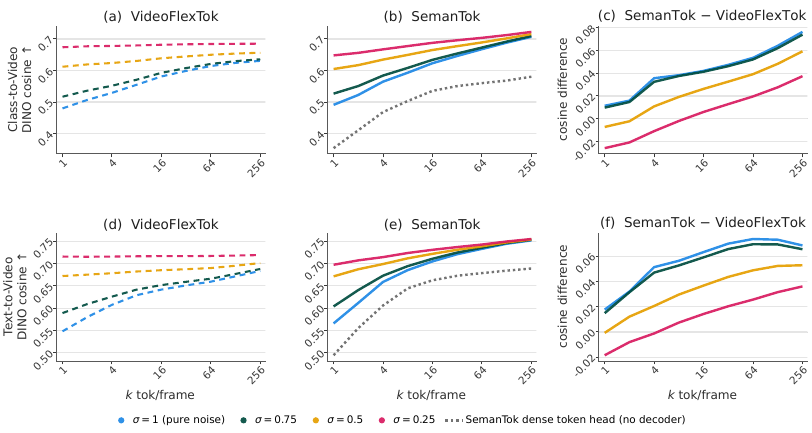}
\vspace{-\parskip}
\vspace{-2.0\abovecaptionskip}
\caption{Where the tokenizers keep DINO semantics.
Decoder-REPA readout of DINOv2 from the first $k$ tokens per frame, 256
validation clips per dataset. Rows: Class-to-Video (Kinetics-600) and
Text-to-Video (uCO3D). Colour: decoder noise level $\sigma$; at $\sigma{=}1$
the input is pure noise, so the prefix is the decoder's only source.
(a,\,d)~\vft{}. (b,\,e)~\ours{}; dotted: its dense token head,
which reads the prefix without the decoder. (c,\,f)~\ours{} minus
\vft{} at each $\sigma$; positive favors \ours{}.
\Cref{fig:dino-probe-main} overlays (a,\,b) and (d,\,e) and adds the gain from $\sigma{=}1$ to $\sigma{=}0.25$.}
\label{fig:dino-probe}
\end{figure*}

\paragraph{At low noise, VideoFlexTok's decoder takes its semantics from the latent.}
At $\sigma{=}0.25$, its readout barely depends on $k$
(\cref{fig:dino-probe}a,\,d).
It moves from $0.716$ at $k{=}1$ to $0.719$ at $k{=}256$ on uCO3D, and
from $0.674$ to $0.685$ on Kinetics-600.
This suggests that the noised latent meets much of its decoder REPA target, which weakens the pressure on the prefix to hold more semantics.
Its curves rise with $k$ mainly at high noise and stay separated by noise
level up to $k{=}256$.
From pure noise its readout rises from $0.548$ to $0.683$ on uCO3D and
from $0.480$ to $0.631$ on Kinetics-600, so its prefix does carry
semantics.
SemanTok's $\sigma{=}0.25$ readout keeps rising with $k$, from $0.697$ to
$0.755$ and from $0.648$ to $0.722$.

\paragraph{SemanTok's prefix carries them.}
From pure noise, SemanTok's readout reaches $0.752$ vs.\ $0.683$ at
$k{=}256$ on uCO3D and $0.707$ vs.\ $0.631$ on Kinetics-600.
Its four noise levels converge as $k$ grows (\cref{fig:dino-probe}b,\,e).
At $k{=}256$ the gap between $\sigma{=}0.25$ and pure noise falls to
$0.003$ on uCO3D ($0.016$ on Kinetics-600), against $0.035$ ($0.054$) for
VideoFlexTok.
SemanTok's prefix alone therefore gives the decoder, by
$k{=}256$, nearly all the DINO
content that the latent adds.
From pure noise, SemanTok reaches VideoFlexTok with a half-clean
latent ($\sigma{=}0.5$) from $k{=}8$ on uCO3D ($0.685$ vs.\ $0.682$) and
$k{=}32$ on Kinetics-600 ($0.646$ vs.\ $0.645$), and with a $75\%$-clean
latent ($\sigma{=}0.25$) from $k{=}32$ ($0.721$ vs.\ $0.716$) and $k{=}128$
($0.688$ vs.\ $0.685$), always at the same $k$.
Decoder REPA is present in both tokenizers, so the difference comes from
the token-side changes.
SemanTok's lead grows with the noise level (\cref{fig:dino-probe}c,\,f).
At $\sigma\ge0.75$ it leads at every $k$ on both datasets, by $0.010$--$0.018$
at $k{=}1$ and $0.065$--$0.076$ at $k{=}256$.
SemanTok's dense token head reads the prefix without the decoder.
It trails every SemanTok decoder readout but also rises monotonically with $k$.
On uCO3D from $k{=}8$ on this two-layer head even exceeds
VideoFlexTok's pure-noise decoder readout ($0.662$ vs.\ $0.641$ at
$k{=}16$).

\paragraph{Limits.}
At the smallest budgets the tokenizers are on par.
The pure-noise difference is at most $0.02$ at $k{=}1$.
With a cleaner latent, VideoFlexTok is ahead up to $k{=}2$ at $\sigma{=}0.5$
and up to $k{=}8$ at $\sigma{=}0.25$, by at most $0.026$.
This matches the mixed $k{\le}4$ regime in \cref{sec:results}.
The probe also does not separate the token-side changes from one another: DINO in the
encoder input, the class-token injection, and the prefix DINO targets.

\section{Experimental details}
\label{sec:app-hparams}

\subsection{Tokenizer training}
\label{sec:app-toktrain}
VideoFlexTok and SemanTok share the tokenizer architecture and optimizer in
\cref{tab:tok-hparams}. Both use a frozen VidTok-128 VAE, an 18-layer
encoder, an 18-layer time-causal decoder, and a six-dimensional FSQ
bottleneck. Nested dropout samples uniformly from powers of two through
$k{=}256$. The decoder REPA weight is 1.0 in both arms.

SemanTok additionally concatenates frozen DINOv2-L features to the encoder
input with a learned linear projection. Its dense and class-token prediction
losses each have weight 0.5. The DINO features are
$16\times16\times1024$ per latent frame, and the class token is projected
into register token zero with a zero-initialized layer.
These heads are causal over latent frames. The paper results use tokenizers
trained on 66B tokens on uCO3D and 131B tokens on Kinetics-600.

\paragraph{Ablations.} Early in the project, we ran smaller-scale ablations to
narrow down the design.
They compared SigLIP~2~\citep{tschannen2025siglip2} with DINOv2 as the
semantic teacher, and we kept DINOv2.
They also favoured using both forms of semantic supervision together: frozen
DINOv2 features at the encoder input, as dense patch features and as a
class-token bias on the first register token, and explicit DINO targets on
every retained token prefix.
\paragraph{DINO frames.} The VAE compresses time causally by $4\times$,
so the 17 RGB frames map to the five latent frames as $\{0\}$,
$\{1..4\}$, $\{5..8\}$, $\{9..12\}$, and $\{13..16\}$.
SemanTok's dense DINO features for a latent frame, used as encoder input
and as the dense target, average DINOv2 over that frame's RGB group.
The first latent frame uses frame 0 alone.
The class token $\mathbf c_t$ comes from the first RGB frame of each group
(frames 0, 1, 5, 9, and 13).

\begin{table}[h]
\centering
\footnotesize
\setlength{\tabcolsep}{4pt}
\begin{tabular}{lll}
\toprule
Knob & uCO3D & Kinetics-600 \\
\midrule
Tokenizer training tokens & 66B & 131B \\
RGB frames / latent frames & 17 / 5 & 17 / 5 \\
Latent grid / channels & $16{\times}16$ / 16 & same \\
Encoder / decoder width & 1152 / 1152 & same \\
Tokens per latent frame & 256 & 256 \\
FSQ levels / vocabulary & $[8,8,8,5,5,5]$ / 64k & same \\
Global / micro batch & 512 / 8 & 512 / 8 \\
Peak LR / warmup tokens & $2.5{\times}10^{-4}$ / 1.3B & $2.5{\times}10^{-4}$ / 3.9B \\
AdamW $\beta$ / grad clip & $(0.9,0.99)$ / 1.0 & same \\
Random-resized-crop scale & $[0.6,1.0]$ & $[0.08,1.0]$ \\
Decoder eval & 50 steps, CFG 3.0 & same \\
\bottomrule
\end{tabular}
\caption{Tokenizer settings used by the reported checkpoints.}
\label{tab:tok-hparams}
\end{table}

\subsection{Tokenizer training budget}
\label{sec:app-tokbudget}
\begin{figure}[h]
\centering
\includegraphics[width=\textwidth]{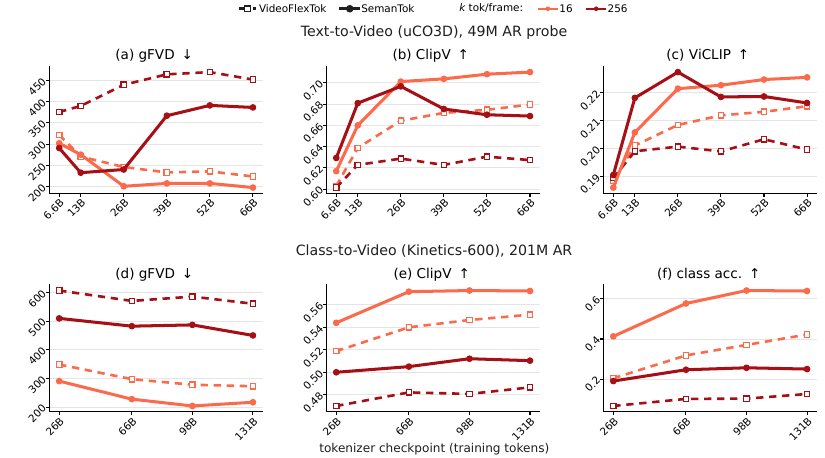}
\vspace{-\parskip}\vspace{-2.0\abovecaptionskip}
\caption{{\ours{}'s lead holds throughout tokenizer training on both datasets.
Each point is an AR model trained on one tokenizer checkpoint's tokens.
Top: uCO3D, a 49M AR probe trained for 8k steps, official validation split.
Bottom: Kinetics-600, a 201M AR model with the recipe and evaluation of \cref{fig:scaling}.
No gap closes as the tokenizer trains longer.}}
\label{fig:tokstep-ar}
\end{figure}

The released VideoFlexTok tokenizer uses roughly 400B training tokens on
Kinetics-600. We train both arms for 131B tokens on Kinetics-600 and 66B on
uCO3D, less than a third of that budget, and our Kinetics-600 tokenizers were
still improving.
AR models trained on earlier tokenizer checkpoints show \ours{}'s lead at
every checkpoint, with no sign of the gap closing (\cref{fig:tokstep-ar}), so
a longer budget is unlikely to reverse it.

\paragraph{Kinetics-600 still improves.} From 66B to 131B tokens, both
tokenizers reconstruct better at $k{=}256$. These numbers score 2{,}560
clips against their VAE-decoded ground truth, not \cref{tab:recon}'s
2{,}048-clip real reference bank, so absolute values and the $k{=}256$
rFVD ordering differ from that table.
VideoFlexTok's rFVD falls from $78.4$ to $59.2$, PSNR rises from $19.0$ to
$19.7$\,dB, and class accuracy from $0.418$ to $0.491$.
SemanTok's rFVD falls from $71.0$ to $54.9$, PSNR rises from $18.1$ to
$18.5$\,dB, and class accuracy from $0.466$ to $0.516$.
We therefore use the 131B checkpoints, and a longer budget would likely
improve both arms further.

\paragraph{uCO3D overfits after 66B.} We also continued both uCO3D
tokenizers to 98B tokens. Reconstruction at $k{=}16$ got worse for both
(256 clips, $k{=}16$): PSNR fell from $18.66$ to $18.07$\,dB for
VideoFlexTok and from $15.96$ to $15.56$\,dB for SemanTok, and rFVD rose
from $351$ to $373$ and from $425$ to $469$.
For a 49M AR model at $k{=}16$, SemanTok's class accuracy and ClipV also
dipped slightly ($0.309\to0.302$ and $0.731\to0.725$), although its gFVD
improved ($226\to205$).
We attribute the decline to overfitting on the small uCO3D training set
and report the 66B checkpoints for both arms.

\paragraph{The uCO3D lead holds throughout training.}
To test whether SemanTok only converges faster, we took six tokenizer
checkpoints between 6.6B and 66B training tokens and trained a fresh 49M
AR model on each one's tokens for 8k steps (\cref{fig:tokstep-ar}).
This recipe is cheaper than \cref{sec:results}'s (unaugmented tokens,
fewer AR steps), so compare only within the figure.
SemanTok has higher ClipV at every checkpoint and both $k$. At 66B tokens
it scores \textbf{0.710 vs.\ 0.680} at $k{=}16$ and \textbf{0.669 vs.\
0.627} at $k{=}256$.
Its gFVD is lower at every checkpoint at $k{=}256$ and from 26B tokens at
$k{=}16$, reaching \textbf{198 vs.\ 224} and \textbf{386 vs.\ 452} at 66B.
ViCLIP and class accuracy show the same ordering (class accuracy
$0.295$ vs.\ $0.214$ at $k{=}16$, 66B).
VideoFlexTok gains little after 26B tokens and closes none of these gaps.
At $k{=}256$ SemanTok's own gFVD rises after 26B, so its lead there
narrows from $199$ to $66$.

\paragraph{The Kinetics-600 lead holds throughout training.}
We repeat this test on Kinetics-600 with the full recipe of \cref{sec:results}
rather than a cheap probe: for tokenizer checkpoints at 26B, 66B, 98B, and
131B training tokens we train a 201M AR model for 20k steps on each one's
tokens and score it with the protocol of \cref{fig:scaling}
(\cref{fig:tokstep-ar}, bottom row).
\ours{} has higher \classacc{} and \ClipV{} at every checkpoint and every
$k$, and lower \gFVD{} at every checkpoint from $k{=}8$.
The gaps do not close with training: at $k{=}16$, \gFVD{} is
\textbf{291 vs.\ 349} at 26B and \textbf{217 vs.\ 273} at 131B, and
\classacc{} is \textbf{0.413 vs.\ 0.205} and \textbf{0.639 vs.\ 0.422}.
At 66B tokens, half the budget, \ours{} already beats \vft{} at 131B on
all three metrics at every $k{\ge}4$, so its lead is not an artifact of the
131B budget.

\paragraph{Released checkpoint.} We cannot compare against the released
VideoFlexTok Kinetics-600 checkpoint. Its decoder is fine-tuned for
bidirectional attention, whereas the paper describes such fine-tuning only
for Panda-70M~\citep{chen2024panda}. The reconstruction metrics we
measured for this checkpoint do not match those reported in the paper,
likely because of this deviation. No time-causal checkpoint is released, so
we retrain VideoFlexTok with the same time-causal recipe, data, and budget
as SemanTok.

\subsection{Autoregressive training}
The downstream AR model is a LLaMA-style causal decoder with RMSNorm and SwiGLU.
At depth $d$, its width is $64d$ and it has $d$ attention heads. It trains
on the time-first sequence of $TK{=}1{,}280$ scalar indices
$\tau_{t,i}$ with global batch 512. A budget $k$ corresponds to the first
$Tk$ scalar indices, or $k$ token positions per latent frame. All runs use
AdamW with $\beta=(0.9,0.95)$, weight decay 0.05, gradient clipping 1.0,
bf16, a 2.5\% warmup, and cosine decay to one percent of peak LR. Head bias
uses the log-unigram initialization. Depth-scaled initialization is disabled.
We otherwise follow VideoFlexTok. For uCO3D only, we increase trunk dropout
and conditioning dropout after observing overfitting.

\begin{table*}[t]
\centering
\footnotesize
\setlength{\tabcolsep}{5pt}
\begin{tabular}{rrrrrr}
\toprule
Non-emb.\ params & Depth & Width & Heads &
uCO3D peak LR & K600 peak LR \\
\midrule
49M   & 10 & 640  & 10 & $8.00{\times}10^{-4}$ & $1.60{\times}10^{-3}$ \\
85M   & 12 & 768  & 12 & $6.67{\times}10^{-4}$ & $1.33{\times}10^{-3}$ \\
201M  & 16 & 1024 & 16 & $5.00{\times}10^{-4}$ & $1.00{\times}10^{-3}$ \\
393M  & 20 & 1280 & 20 & $4.00{\times}10^{-4}$ & $8.00{\times}10^{-4}$ \\
679M  & 24 & 1536 & 24 & $3.33{\times}10^{-4}$ & $6.67{\times}10^{-4}$ \\
1.33B & 30 & 1920 & 30 & $2.67{\times}10^{-4}$ & $5.33{\times}10^{-4}$ \\
2.29B & 36 & 2304 & 36 & $2.22{\times}10^{-4}$ & $4.44{\times}10^{-4}$ \\
\bottomrule
\end{tabular}
\caption{AR model size ladder. uCO3D uses $0.512/\text{width}$; Kinetics-600
uses the VideoFlexTok rule $1.024/\text{width}$.}
\label{tab:ar-size-hparams}
\end{table*}

\begin{table}[h]
\centering
\footnotesize
\setlength{\tabcolsep}{4pt}
\begin{tabular}{ll}
\toprule
Knob & Dataset-specific value \\
\midrule
Conditioning & uCO3D: umT5, 128 tokens; K600: class \\
Tokenizer crop views & uCO3D: 25; K600: 6 \\
AR training tokens & uCO3D: 13.1B at every size \\
& K600: 13.1B through 679M; 26.2B thereafter; 65.5B for \cref{fig:trainD} \\
Trunk dropout & uCO3D: 0.25; K600: 0.1 \\
Condition drop & uCO3D: 0.2; K600: 0.1 \\
Condition-feature drop & uCO3D: 0.1; K600: 0 \\
Cross-attention drop & 0 \\
\bottomrule
\end{tabular}
\caption{Dataset-specific AR settings. Crop views affect only AR training;
validation tokens are unaugmented.}
\label{tab:ar-data-hparams}
\end{table}

\subsection{Evaluation protocols}
\paragraph{AR generation.}
We follow the VideoFlexTok evaluation pipeline. Sampling uses temperature
1.0 without top-$k$ or top-$p$ truncation. The decoder uses 50 flow steps
and guidance 3.0. Kinetics-600 AR guidance is 3.0 for
$k\in\{1,4\}$, 2.0 for $k\in\{8,16,32\}$, and 1.0 thereafter,
following the budget dependence reported by VideoFlexTok. uCO3D uses
AR guidance 3.0 at every budget. We apply the same settings to both
tokenizers and did not tune guidance for either.

\paragraph{Sample sizes and budget selection.} On Kinetics-600, each
$(\text{AR size}, k)$ cell uses 2{,}048 generated clips, conditioned on
labels drawn from 2{,}048 validation clips that also form the real FVD and
FID reference. On uCO3D, each split (ID and OOD) uses 2{,}560 generated
clips against 1{,}024 real clips. Figure summaries weight the splits
1{,}014:152, as in the official validation set; gFVD and gFID are pooled
Fr\'echet distances, and the other metrics are weighted means. Table-level
ID/OOD numbers (\cref{tab:catacc}) are per split. All numbers come from a
single training run and sampling seed; \cref{sec:app-ci} gives bootstrap
intervals for the headline comparisons. Best-$k$ values are selected per
metric on the same evaluation set, which favors both arms equally; the
fixed-$k$ comparisons in \cref{sec:results} involve no selection.

\paragraph{Tokenizer reconstruction.}
\Cref{tab:recon} contains no AR-model rollouts. We decode ground-truth tokens
and apply the same rFVD, ViCLIP, ClipV, class accuracy, PSNR, and SSIM evaluation
used for generation.

\paragraph{Semantic alignment.}
ViCLIP scores generated video against the conditioning caption. ClipV
scores the same video against the VAE-decoded ground-truth clip in the
ViCLIP video encoder, with no text encoding.

\paragraph{Class accuracy.}
On Kinetics-600, a UMT-L classifier finetuned for Kinetics-600 scores the
first 16 of 17 frames, resized to $224{\times}224$, as in
VideoFlexTok~\citep{videoflextok,li2023umt}.
Top-1 is taken against the clip's action class.
For AR generation that class is the conditioning label.
On uCO3D, each class centroid is the L2-normalized mean of real-frame
Inception features in the evaluation pool.
A frame is correct if its nearest centroid, by cosine, is the clip's true
object class.
NCM stays defined for OOD classes, which a closed-set classifier never saw.
This readout is coarse, so on uCO3D we also report \ViCLIP{} and \ClipV{}.
Tokenizer reconstruction uses the same two readouts on decoded
ground-truth tokens.

\paragraph{Cross-entropy and teacher forcing.}
The cross-entropy probe uses ground-truth previous tokens and reports
bits/token. In the hybrid Kinetics-600 experiment, we teacher-force $m$
ground-truth token positions per frame, supplying $Tm$ scalar indices,
and then free-run to $k{=}256$ with the same AR model and decoder sampling
settings.
Cross-entropy is measured on 1{,}024 validation clips. The marginal
entropy $H$ of a position is the plug-in entropy of its code, with the
Miller--Madow correction, over 120k training clips; context bits are $H$
minus cross-entropy.
Prefix bits per frame sum the per-position cross-entropy over the first
$k$ positions.

\begin{figure}[h]
\centering
\includegraphics[width=0.54\textwidth]{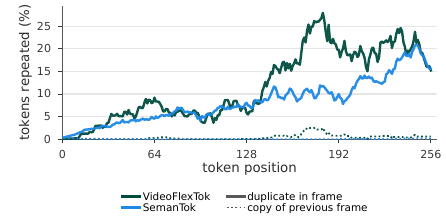}
\caption{Token repeats by position on Kinetics-600 (running mean over 16
positions). Solid: duplicates of an earlier token in the same frame.
Dotted: copies of the previous frame's token at the same position.}
\label{fig:repeats}
\end{figure}

\paragraph{Token repeats.}
\label{sec:app-repeats}
On 60k Kinetics-600 training clips, a \emph{duplicate} is a token equal to
any earlier token of its frame, and a \emph{copy} is a token equal to the
one at the same position in the previous latent frame (\cref{fig:repeats}).
Over all 256 positions, SemanTok duplicates $8.5\%$ of tokens against
VideoFlexTok's $12.0\%$, and copies $0.05\%$ against $0.42\%$.
In the first 16 positions, where its cross-entropy advantage is largest,
both tokenizers stay near zero (SemanTok: $1.1\%$ duplicates and $0.14\%$
copies; VideoFlexTok: $0.2\%$ and $0.02\%$).

\subsection{Uncertainty of the generation comparisons}
\label{sec:app-ci}

We estimate uncertainty with a paired bootstrap over the evaluation clips,
using 1{,}000 replicates.
Both tokenizers condition on the same prompts or labels, so each replicate
resamples the two arms jointly.
We report SemanTok's advantage, signed so that positive values favour
SemanTok, with its 95\% interval.
At $k{=}16$ we also pool six sampling seeds, which change only the AR and
flow noise.
On uCO3D the ID validation split holds only 1{,}014 clips, which caps the
real reference.

\begin{figure}[t]
\centering
\includegraphics[width=\textwidth]{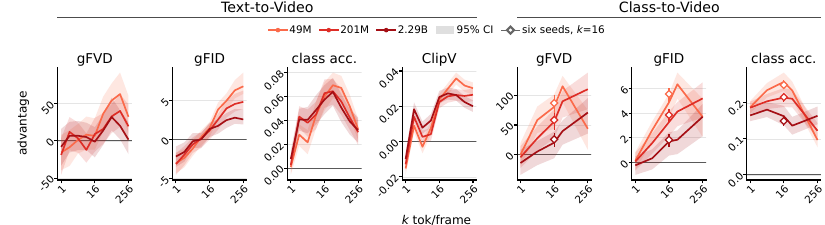}
\caption{SemanTok's advantage over VideoFlexTok at equal AR size, as
a function of the token budget. Each panel shows the paired difference
between the two tokenizers under the same AR size, $k$, prompts, and
labels, signed so that positive values favour SemanTok (VideoFlexTok minus
SemanTok for gFVD and gFID; SemanTok minus VideoFlexTok for class
accuracy and ClipV). Lines are point estimates for three AR sizes; bands
are 95\% bootstrap intervals over evaluation clips; the grey line marks no
difference. SemanTok's semantic-alignment lead appears from $k{=}2$, its fidelity
lead from $k{\approx}16$ on uCO3D and from $k{=}4$ (small AR models) to
$k{=}32$ (largest) on Kinetics-600, and
VideoFlexTok is ahead only at $k{\le}4$ on uCO3D. Diamonds: Kinetics-600
at $k{=}16$ with six sampling seeds pooled.}
\label{fig:ci-strip}
\end{figure}

\begin{figure}[p]
\centering
\includegraphics[width=\textwidth]{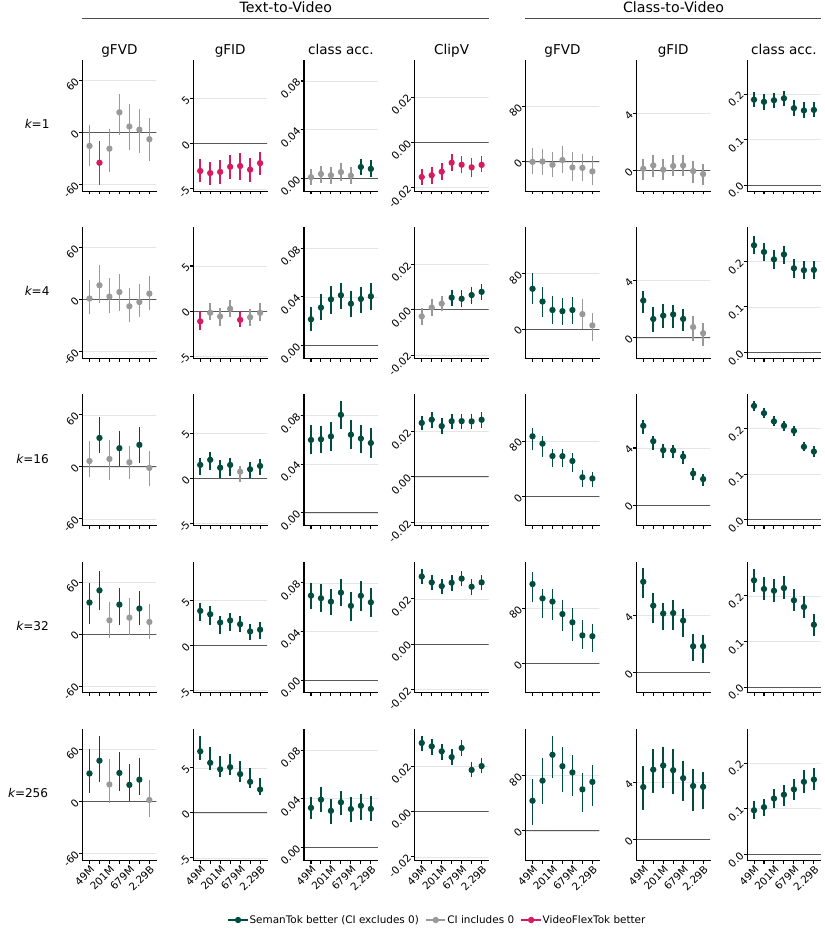}
\caption{SemanTok's advantage for every AR size at five token
budgets. Rows fix $k$, columns fix the metric, and within each panel the
AR size grows from 49M (left) to 2.29B (right). Dots are the paired
difference at equal AR size and $k$, signed as in \cref{fig:ci-strip} so
that positive values favour SemanTok; whiskers are 95\% bootstrap
intervals over evaluation clips. Teal marks an interval above zero, grey an
interval that includes zero, and magenta an interval below zero. Each
column shares its $y$-range, so the change with $k$ reads down the page.
The Kinetics-600 $k{=}16$ row pools six sampling seeds; all other cells use
one.}
\label{fig:ci-ladder}
\end{figure}

\paragraph{Equal AR size.}
\Cref{fig:ci-strip,fig:ci-ladder} compare the tokenizers at equal AR size
and budget.
Semantic alignment favours SemanTok almost everywhere, with class accuracy from
$k{=}2$ on uCO3D and at every budget on Kinetics-600, and ClipV from
$k{=}8$.
Fidelity needs longer prefixes.
On uCO3D, gFID gains exclude zero at every size from $k{=}32$, and gFVD
gains at $k\in\{64,128\}$.
On Kinetics-600, with six seeds pooled at $k{=}16$, gFVD, gFID, and class
accuracy exclude zero at every size.
VideoFlexTok is significantly ahead only at $k{\le}4$ on uCO3D, mostly at
$k{=}1$.
With each arm at its own best $k$, every tested metric excludes zero at
every size except uCO3D gFVD, which does so at two of seven sizes.

\paragraph{Across AR sizes.}
Against larger VideoFlexTok AR models, the small SemanTok AR model's semantic-alignment lead excludes zero in every comparison, for example $+0.053$ class accuracy for
201M vs.\ 679M on uCO3D.
With six seeds pooled, its gFID lead does too, for example $+1.19$
$[0.81, 1.55]$ for 201M vs.\ 679M on uCO3D, where the pooled estimate also uses
every generated clip.
On Kinetics-600, the 85M SemanTok AR model's gFVD advantage over the 1.33B
VideoFlexTok AR model also excludes zero ($+19.1$ $[7.5, 28.9]$).
Only uCO3D gFVD remains at parity, for example $+3.1$ $[-3.1, 9.0]$ for
201M vs.\ 679M.

\section{Generation and decoder REPA per AR size and budget}
\label{sec:app-gen-tables}

\begin{table}[p]
\caption{Kinetics-600 class-to-video generation: fidelity. AR models sampled with $k$ tokens per frame, tokenizers at 200k steps, AR models at 20k steps (1.33B and 2.29B at 40k), $n{=}2048$. Bold is better at that $k$ and size; $^\dagger$ marks the better tokenizer where the 95\% paired-bootstrap CI of the difference (single sampling seed) excludes zero. CIs exist at $k\in\{1,4,16,32,256\}$ only. $^\ddagger$ marks \ours{} gains that exclude zero once six sampling seeds are pooled; with pooling, every $k{=}16$ gFVD and gFID gain of \ours{} excludes zero.}
\label{tab:gen-kinetics}
\centering
\footnotesize
\setlength{\tabcolsep}{3.2pt}
\begin{tabular}{@{}ll *{8}{c}@{}}
\toprule
AR & & $k{=}1$ & $4$ & $8$ & $16$ & $32$ & $64$ & $128$ & $256$ \\
\midrule
\multicolumn{10}{@{}l}{\textbf{gFVD}$\downarrow$} \\
\multirow{2}{*}{49M} & \vft{} & $390.8$ & $295.7$ & $317.5$ & $303.9$ & $338.8$ & $478.6$ & $588.9$ & $653.7$ \\
 & \ours{} & $\mathbf{390.4}$ & $\mathbf{236.8}^{\dagger}$ & $\mathbf{239.2}$ & $\mathbf{223.7}^{\dagger}$ & $\mathbf{223.7}^{\dagger}$ & $\mathbf{324.0}$ & $\mathbf{450.2}$ & $\mathbf{610.1}^{\dagger}$ \\[1pt]
\multirow{2}{*}{85M} & \vft{} & $386.3$ & $284.1$ & $293.3$ & $278.9$ & $307.0$ & $450.6$ & $548.9$ & $627.3$ \\
 & \ours{} & $\mathbf{385.4}$ & $\mathbf{244.0}^{\dagger}$ & $\mathbf{228.8}$ & $\mathbf{219.9}^{\dagger}$ & $\mathbf{212.5}^{\dagger}$ & $\mathbf{322.8}$ & $\mathbf{411.0}$ & $\mathbf{554.5}^{\dagger}$ \\[1pt]
\multirow{2}{*}{201M} & \vft{} & $\mathbf{384.4}$ & $278.9$ & $279.9$ & $272.9$ & $302.9$ & $420.7$ & $501.0$ & $560.3$ \\
 & \ours{} & $388.3$ & $\mathbf{251.0}^{\dagger}$ & $\mathbf{240.8}$ & $\mathbf{217.2}^{\dagger}$ & $\mathbf{213.0}^{\dagger}$ & $\mathbf{295.2}$ & $\mathbf{361.2}$ & $\mathbf{449.8}^{\dagger}$ \\[1pt]
\multirow{2}{*}{393M} & \vft{} & $394.9$ & $274.1$ & $273.0$ & $263.8$ & $284.3$ & $400.7$ & $475.1$ & $521.6$ \\
 & \ours{} & $\mathbf{392.1}$ & $\mathbf{247.9}^{\dagger}$ & $\mathbf{233.4}$ & $\mathbf{212.7}^{\dagger}$ & $\mathbf{212.6}^{\dagger}$ & $\mathbf{286.7}$ & $\mathbf{348.0}$ & $\mathbf{428.0}^{\dagger}$ \\[1pt]
\multirow{2}{*}{679M} & \vft{} & $\mathbf{380.5}$ & $271.8$ & $264.1$ & $249.1$ & $267.0$ & $390.6$ & $451.1$ & $475.6$ \\
 & \ours{} & $388.1$ & $\mathbf{243.9}^{\dagger}$ & $\mathbf{222.4}$ & $\mathbf{216.5}^{\dagger}$ & $\mathbf{207.1}^{\dagger}$ & $\mathbf{279.9}$ & $\mathbf{332.5}$ & $\mathbf{390.7}^{\dagger}$ \\[1pt]
\multirow{2}{*}{1.33B} & \vft{} & $\mathbf{388.4}$ & $266.0$ & $257.9$ & $222.8$ & $239.5$ & $337.6$ & $382.2$ & $391.6$ \\
 & \ours{} & $397.0$ & $\mathbf{243.9}$ & $\mathbf{229.0}$ & $\mathbf{211.3}^{\ddagger}$ & $\mathbf{198.3}^{\dagger}$ & $\mathbf{256.9}$ & $\mathbf{277.6}$ & $\mathbf{331.7}^{\dagger}$ \\[1pt]
\multirow{2}{*}{2.29B} & \vft{} & $\mathbf{389.2}$ & $255.6$ & $245.2$ & $225.9$ & $241.5$ & $330.9$ & $367.7$ & $383.2$ \\
 & \ours{} & $402.6$ & $\mathbf{249.8}$ & $\mathbf{237.5}$ & $\mathbf{205.7}^{\ddagger}$ & $\mathbf{201.7}^{\dagger}$ & $\mathbf{258.2}$ & $\mathbf{278.9}$ & $\mathbf{312.2}^{\dagger}$ \\
\midrule
\multicolumn{10}{@{}l}{\textbf{gFID}$\downarrow$} \\
\multirow{2}{*}{49M} & \vft{} & $32.90$ & $26.87$ & $25.51$ & $26.16$ & $28.04$ & $33.97$ & $40.87$ & $46.54$ \\
 & \ours{} & $\mathbf{32.76}$ & $\mathbf{24.27}^{\dagger}$ & $\mathbf{22.93}$ & $\mathbf{21.31}^{\dagger}$ & $\mathbf{21.71}^{\dagger}$ & $\mathbf{26.26}$ & $\mathbf{32.94}$ & $\mathbf{42.83}^{\dagger}$ \\[1pt]
\multirow{2}{*}{85M} & \vft{} & $32.89$ & $26.37$ & $24.89$ & $25.17$ & $26.54$ & $32.51$ & $38.79$ & $44.80$ \\
 & \ours{} & $\mathbf{32.52}$ & $\mathbf{25.05}^{\dagger}$ & $\mathbf{22.77}$ & $\mathbf{21.58}^{\dagger}$ & $\mathbf{21.88}^{\dagger}$ & $\mathbf{25.99}$ & $\mathbf{31.51}$ & $\mathbf{39.85}^{\dagger}$ \\[1pt]
\multirow{2}{*}{201M} & \vft{} & $32.73$ & $26.16$ & $24.86$ & $24.51$ & $25.49$ & $31.44$ & $36.80$ & $41.14$ \\
 & \ours{} & $\mathbf{32.67}$ & $\mathbf{24.61}^{\dagger}$ & $\mathbf{23.56}$ & $\mathbf{21.26}^{\dagger}$ & $\mathbf{21.38}^{\dagger}$ & $\mathbf{24.92}$ & $\mathbf{29.36}$ & $\mathbf{35.92}^{\dagger}$ \\[1pt]
\multirow{2}{*}{393M} & \vft{} & $33.11$ & $25.98$ & $24.55$ & $24.92$ & $25.60$ & $30.88$ & $35.51$ & $39.14$ \\
 & \ours{} & $\mathbf{32.77}$ & $\mathbf{24.34}^{\dagger}$ & $\mathbf{23.09}$ & $\mathbf{21.08}^{\dagger}$ & $\mathbf{21.45}^{\dagger}$ & $\mathbf{24.85}$ & $\mathbf{28.43}$ & $\mathbf{34.24}^{\dagger}$ \\[1pt]
\multirow{2}{*}{679M} & \vft{} & $32.91$ & $25.71$ & $23.47$ & $23.16$ & $24.73$ & $30.54$ & $34.09$ & $37.37$ \\
 & \ours{} & $\mathbf{32.55}$ & $\mathbf{24.39}^{\dagger}$ & $\mathbf{23.16}$ & $\mathbf{21.23}^{\dagger}$ & $\mathbf{21.13}^{\dagger}$ & $\mathbf{24.38}$ & $\mathbf{27.49}$ & $\mathbf{33.05}^{\dagger}$ \\[1pt]
\multirow{2}{*}{1.33B} & \vft{} & $\mathbf{32.70}$ & $24.59$ & $\mathbf{22.92}$ & $21.94$ & $22.61$ & $26.50$ & $30.92$ & $32.33$ \\
 & \ours{} & $32.74$ & $\mathbf{23.85}$ & $22.97$ & $\mathbf{21.15}^{\ddagger}$ & $\mathbf{20.80}^{\dagger}$ & $\mathbf{23.96}$ & $\mathbf{24.85}$ & $\mathbf{28.55}^{\dagger}$ \\[1pt]
\multirow{2}{*}{2.29B} & \vft{} & $\mathbf{32.45}$ & $24.09$ & $\mathbf{22.25}$ & $22.37$ & $22.81$ & $26.61$ & $28.99$ & $30.94$ \\
 & \ours{} & $32.68$ & $\mathbf{23.77}$ & $23.45$ & $\mathbf{20.69}^{\dagger}$ & $\mathbf{21.01}^{\dagger}$ & $\mathbf{23.66}$ & $\mathbf{24.67}$ & $\mathbf{27.21}^{\dagger}$ \\
\bottomrule
\end{tabular}
\end{table}

\begin{table}[p]
\caption{Kinetics-600 class-to-video generation: semantic alignment. Same samples as \Cref{tab:gen-kinetics}. Class acc.\ is UMT-L top-1. Bold is better at that $k$ and size; $^\dagger$ marks the better tokenizer where the 95\% paired-bootstrap CI of the difference (single sampling seed) excludes zero. CIs exist for class acc.\ at $k\in\{1,4,16,32,256\}$ only; ViCLIP and ClipV have none.}
\label{tab:gen-kinetics-sem}
\centering
\footnotesize
\setlength{\tabcolsep}{3.2pt}
\begin{tabular}{@{}ll *{8}{c}@{}}
\toprule
AR & & $k{=}1$ & $4$ & $8$ & $16$ & $32$ & $64$ & $128$ & $256$ \\
\midrule
\multicolumn{10}{@{}l}{\textbf{ViCLIP}$\uparrow$} \\
\multirow{2}{*}{49M} & \vft{} & $0.179$ & $0.192$ & $0.187$ & $0.189$ & $0.187$ & $0.161$ & $0.160$ & $0.156$ \\
 & \ours{} & $\mathbf{0.193}$ & $\mathbf{0.208}$ & $\mathbf{0.206}$ & $\mathbf{0.208}$ & $\mathbf{0.206}$ & $\mathbf{0.179}$ & $\mathbf{0.175}$ & $\mathbf{0.168}$ \\[1pt]
\multirow{2}{*}{85M} & \vft{} & $0.178$ & $0.193$ & $0.188$ & $0.191$ & $0.190$ & $0.164$ & $0.161$ & $0.159$ \\
 & \ours{} & $\mathbf{0.194}$ & $\mathbf{0.208}$ & $\mathbf{0.208}$ & $\mathbf{0.208}$ & $\mathbf{0.206}$ & $\mathbf{0.181}$ & $\mathbf{0.177}$ & $\mathbf{0.171}$ \\[1pt]
\multirow{2}{*}{201M} & \vft{} & $0.178$ & $0.194$ & $0.191$ & $0.194$ & $0.192$ & $0.165$ & $0.163$ & $0.161$ \\
 & \ours{} & $\mathbf{0.193}$ & $\mathbf{0.207}$ & $\mathbf{0.207}$ & $\mathbf{0.208}$ & $\mathbf{0.208}$ & $\mathbf{0.183}$ & $\mathbf{0.181}$ & $\mathbf{0.176}$ \\[1pt]
\multirow{2}{*}{393M} & \vft{} & $0.178$ & $0.194$ & $0.190$ & $0.194$ & $0.194$ & $0.166$ & $0.165$ & $0.163$ \\
 & \ours{} & $\mathbf{0.193}$ & $\mathbf{0.208}$ & $\mathbf{0.208}$ & $\mathbf{0.208}$ & $\mathbf{0.209}$ & $\mathbf{0.184}$ & $\mathbf{0.182}$ & $\mathbf{0.177}$ \\[1pt]
\multirow{2}{*}{679M} & \vft{} & $0.178$ & $0.194$ & $0.193$ & $0.196$ & $0.197$ & $0.168$ & $0.168$ & $0.165$ \\
 & \ours{} & $\mathbf{0.192}$ & $\mathbf{0.206}$ & $\mathbf{0.207}$ & $\mathbf{0.209}$ & $\mathbf{0.210}$ & $\mathbf{0.185}$ & $\mathbf{0.184}$ & $\mathbf{0.180}$ \\[1pt]
\multirow{2}{*}{1.33B} & \vft{} & $0.177$ & $0.193$ & $0.193$ & $0.199$ & $0.200$ & $0.172$ & $0.171$ & $0.170$ \\
 & \ours{} & $\mathbf{0.191}$ & $\mathbf{0.205}$ & $\mathbf{0.207}$ & $\mathbf{0.209}$ & $\mathbf{0.209}$ & $\mathbf{0.186}$ & $\mathbf{0.186}$ & $\mathbf{0.184}$ \\[1pt]
\multirow{2}{*}{2.29B} & \vft{} & $0.176$ & $0.192$ & $0.193$ & $0.200$ & $0.201$ & $0.173$ & $0.172$ & $0.172$ \\
 & \ours{} & $\mathbf{0.190}$ & $\mathbf{0.204}$ & $\mathbf{0.205}$ & $\mathbf{0.209}$ & $\mathbf{0.208}$ & $\mathbf{0.187}$ & $\mathbf{0.187}$ & $\mathbf{0.186}$ \\
\midrule
\multicolumn{10}{@{}l}{\textbf{ClipV}$\uparrow$} \\
\multirow{2}{*}{49M} & \vft{} & $0.531$ & $0.550$ & $0.543$ & $0.546$ & $0.539$ & $0.491$ & $0.487$ & $0.475$ \\
 & \ours{} & $\mathbf{0.549}$ & $\mathbf{0.577}$ & $\mathbf{0.574}$ & $\mathbf{0.574}$ & $\mathbf{0.570}$ & $\mathbf{0.522}$ & $\mathbf{0.513}$ & $\mathbf{0.495}$ \\[1pt]
\multirow{2}{*}{85M} & \vft{} & $0.531$ & $0.552$ & $0.545$ & $0.548$ & $0.545$ & $0.493$ & $0.489$ & $0.480$ \\
 & \ours{} & $\mathbf{0.549}$ & $\mathbf{0.575}$ & $\mathbf{0.574}$ & $\mathbf{0.574}$ & $\mathbf{0.571}$ & $\mathbf{0.525}$ & $\mathbf{0.517}$ & $\mathbf{0.501}$ \\[1pt]
\multirow{2}{*}{201M} & \vft{} & $0.529$ & $0.552$ & $0.549$ & $0.551$ & $0.548$ & $0.497$ & $0.492$ & $0.487$ \\
 & \ours{} & $\mathbf{0.546}$ & $\mathbf{0.575}$ & $\mathbf{0.573}$ & $\mathbf{0.572}$ & $\mathbf{0.571}$ & $\mathbf{0.528}$ & $\mathbf{0.521}$ & $\mathbf{0.510}$ \\[1pt]
\multirow{2}{*}{393M} & \vft{} & $0.527$ & $0.551$ & $0.545$ & $0.549$ & $0.550$ & $0.500$ & $0.494$ & $0.491$ \\
 & \ours{} & $\mathbf{0.544}$ & $\mathbf{0.572}$ & $\mathbf{0.571}$ & $\mathbf{0.573}$ & $\mathbf{0.574}$ & $\mathbf{0.532}$ & $\mathbf{0.524}$ & $\mathbf{0.511}$ \\[1pt]
\multirow{2}{*}{679M} & \vft{} & $0.527$ & $0.552$ & $0.550$ & $0.555$ & $0.556$ & $0.504$ & $0.502$ & $0.492$ \\
 & \ours{} & $\mathbf{0.544}$ & $\mathbf{0.569}$ & $\mathbf{0.570}$ & $\mathbf{0.572}$ & $\mathbf{0.575}$ & $\mathbf{0.534}$ & $\mathbf{0.526}$ & $\mathbf{0.516}$ \\[1pt]
\multirow{2}{*}{1.33B} & \vft{} & $0.524$ & $0.548$ & $0.550$ & $0.556$ & $0.561$ & $0.509$ & $0.504$ & $0.503$ \\
 & \ours{} & $\mathbf{0.542}$ & $\mathbf{0.566}$ & $\mathbf{0.571}$ & $\mathbf{0.571}$ & $\mathbf{0.572}$ & $\mathbf{0.533}$ & $\mathbf{0.532}$ & $\mathbf{0.523}$ \\[1pt]
\multirow{2}{*}{2.29B} & \vft{} & $0.523$ & $0.547$ & $0.548$ & $0.557$ & $0.561$ & $0.511$ & $0.509$ & $0.507$ \\
 & \ours{} & $\mathbf{0.539}$ & $\mathbf{0.564}$ & $\mathbf{0.567}$ & $\mathbf{0.571}$ & $\mathbf{0.570}$ & $\mathbf{0.535}$ & $\mathbf{0.534}$ & $\mathbf{0.528}$ \\
\midrule
\multicolumn{10}{@{}l}{\textbf{class acc.}$\uparrow$} \\
\multirow{2}{*}{49M} & \vft{} & $0.234$ & $0.392$ & $0.303$ & $0.340$ & $0.335$ & $0.109$ & $0.115$ & $0.082$ \\
 & \ours{} & $\mathbf{0.424}^{\dagger}$ & $\mathbf{0.631}^{\dagger}$ & $\mathbf{0.589}$ & $\mathbf{0.596}^{\dagger}$ & $\mathbf{0.568}^{\dagger}$ & $\mathbf{0.280}$ & $\mathbf{0.234}$ & $\mathbf{0.180}^{\dagger}$ \\[1pt]
\multirow{2}{*}{85M} & \vft{} & $0.242$ & $0.408$ & $0.321$ & $0.377$ & $0.381$ & $0.126$ & $0.117$ & $0.101$ \\
 & \ours{} & $\mathbf{0.426}^{\dagger}$ & $\mathbf{0.628}^{\dagger}$ & $\mathbf{0.613}$ & $\mathbf{0.618}^{\dagger}$ & $\mathbf{0.593}^{\dagger}$ & $\mathbf{0.300}$ & $\mathbf{0.267}$ & $\mathbf{0.208}^{\dagger}$ \\[1pt]
\multirow{2}{*}{201M} & \vft{} & $0.242$ & $0.436$ & $0.381$ & $0.422$ & $0.420$ & $0.143$ & $0.146$ & $0.126$ \\
 & \ours{} & $\mathbf{0.426}^{\dagger}$ & $\mathbf{0.640}^{\dagger}$ & $\mathbf{0.627}$ & $\mathbf{0.639}^{\dagger}$ & $\mathbf{0.631}^{\dagger}$ & $\mathbf{0.318}$ & $\mathbf{0.303}$ & $\mathbf{0.250}^{\dagger}$ \\[1pt]
\multirow{2}{*}{393M} & \vft{} & $0.244$ & $0.446$ & $0.390$ & $0.444$ & $0.441$ & $0.158$ & $0.155$ & $0.137$ \\
 & \ours{} & $\mathbf{0.433}^{\dagger}$ & $\mathbf{0.659}^{\dagger}$ & $\mathbf{0.629}$ & $\mathbf{0.646}^{\dagger}$ & $\mathbf{0.658}^{\dagger}$ & $\mathbf{0.333}$ & $\mathbf{0.313}$ & $\mathbf{0.267}^{\dagger}$ \\[1pt]
\multirow{2}{*}{679M} & \vft{} & $0.250$ & $0.461$ & $0.421$ & $0.471$ & $0.474$ & $0.165$ & $0.164$ & $0.154$ \\
 & \ours{} & $\mathbf{0.419}^{\dagger}$ & $\mathbf{0.649}^{\dagger}$ & $\mathbf{0.632}$ & $\mathbf{0.673}^{\dagger}$ & $\mathbf{0.665}^{\dagger}$ & $\mathbf{0.363}$ & $\mathbf{0.341}$ & $\mathbf{0.292}^{\dagger}$ \\[1pt]
\multirow{2}{*}{1.33B} & \vft{} & $0.242$ & $0.466$ & $0.448$ & $0.547$ & $0.535$ & $0.204$ & $0.206$ & $0.202$ \\
 & \ours{} & $\mathbf{0.407}^{\dagger}$ & $\mathbf{0.650}^{\dagger}$ & $\mathbf{0.667}$ & $\mathbf{0.701}^{\dagger}$ & $\mathbf{0.710}^{\dagger}$ & $\mathbf{0.408}$ & $\mathbf{0.384}$ & $\mathbf{0.357}^{\dagger}$ \\[1pt]
\multirow{2}{*}{2.29B} & \vft{} & $0.237$ & $0.467$ & $0.453$ & $0.545$ & $0.560$ & $0.235$ & $0.209$ & $0.217$ \\
 & \ours{} & $\mathbf{0.400}^{\dagger}$ & $\mathbf{0.650}^{\dagger}$ & $\mathbf{0.647}$ & $\mathbf{0.701}^{\dagger}$ & $\mathbf{0.700}^{\dagger}$ & $\mathbf{0.401}$ & $\mathbf{0.412}$ & $\mathbf{0.384}^{\dagger}$ \\
\bottomrule
\end{tabular}
\end{table}

\begin{table}[p]
\caption{uCO3D text-to-video generation: fidelity. Tokenizers at 100k steps, AR models at 20k steps; 1{,}014 ID and 152 OOD clips pooled, gFVD and gFID as one Fr\'echet distance. Bold is better at that $k$ and size; $^\dagger$ marks the better tokenizer where the 95\% paired-bootstrap CI of the difference (single sampling seed) excludes zero. CIs cover every cell.}
\label{tab:gen-uco3d}
\centering
\footnotesize
\setlength{\tabcolsep}{3.2pt}
\begin{tabular}{@{}ll *{8}{c}@{}}
\toprule
AR & & $k{=}1$ & $4$ & $8$ & $16$ & $32$ & $64$ & $128$ & $256$ \\
\midrule
\multicolumn{10}{@{}l}{\textbf{gFVD}$\downarrow$} \\
\multirow{2}{*}{49M} & \vft{} & $\mathbf{436.3}$ & $259.9$ & $235.2$ & $232.9$ & $254.7$ & $284.5$ & $344.1$ & $379.1$ \\
 & \ours{} & $452.0$ & $\mathbf{258.7}$ & $\mathbf{226.9}$ & $\mathbf{226.2}$ & $\mathbf{218.2}^{\dagger}$ & $\mathbf{230.4}^{\dagger}$ & $\mathbf{281.2}^{\dagger}$ & $\mathbf{346.9}^{\dagger}$ \\[1pt]
\multirow{2}{*}{85M} & \vft{} & $\mathbf{426.8}^{\dagger}$ & $255.9$ & $\mathbf{229.0}$ & $237.1$ & $260.1$ & $275.5$ & $338.4$ & $367.1$ \\
 & \ours{} & $461.9$ & $\mathbf{239.4}$ & $233.9$ & $\mathbf{203.7}^{\dagger}$ & $\mathbf{209.6}^{\dagger}$ & $\mathbf{218.9}^{\dagger}$ & $\mathbf{261.1}^{\dagger}$ & $\mathbf{320.1}^{\dagger}$ \\[1pt]
\multirow{2}{*}{201M} & \vft{} & $\mathbf{424.4}$ & $244.2$ & $\mathbf{211.8}$ & $218.6$ & $224.5$ & $244.0$ & $286.8$ & $315.5$ \\
 & \ours{} & $443.2$ & $\mathbf{240.7}$ & $223.9$ & $\mathbf{209.5}$ & $\mathbf{208.6}$ & $\mathbf{211.3}^{\dagger}$ & $\mathbf{246.6}^{\dagger}$ & $\mathbf{296.0}$ \\[1pt]
\multirow{2}{*}{393M} & \vft{} & $448.1$ & $246.4$ & $\mathbf{218.5}$ & $224.3$ & $233.8$ & $253.5$ & $288.6$ & $306.0$ \\
 & \ours{} & $\mathbf{425.0}$ & $\mathbf{237.7}$ & $219.8$ & $\mathbf{202.7}^{\dagger}$ & $\mathbf{199.7}^{\dagger}$ & $\mathbf{206.1}^{\dagger}$ & $\mathbf{228.6}^{\dagger}$ & $\mathbf{273.2}^{\dagger}$ \\[1pt]
\multirow{2}{*}{679M} & \vft{} & $435.8$ & $\mathbf{232.2}$ & $\mathbf{208.2}$ & $213.3$ & $221.4$ & $245.4$ & $263.8$ & $288.8$ \\
 & \ours{} & $\mathbf{429.1}$ & $239.9$ & $210.2$ & $\mathbf{208.0}$ & $\mathbf{202.4}$ & $\mathbf{199.1}^{\dagger}$ & $\mathbf{223.6}^{\dagger}$ & $\mathbf{269.7}^{\dagger}$ \\[1pt]
\multirow{2}{*}{1.33B} & \vft{} & $432.2$ & $\mathbf{235.2}$ & $221.3$ & $223.1$ & $221.7$ & $225.4$ & $255.1$ & $273.4$ \\
 & \ours{} & $\mathbf{429.1}$ & $238.0$ & $\mathbf{209.3}$ & $\mathbf{197.7}^{\dagger}$ & $\mathbf{192.1}^{\dagger}$ & $\mathbf{201.6}^{\dagger}$ & $\mathbf{218.6}^{\dagger}$ & $\mathbf{248.3}^{\dagger}$ \\[1pt]
\multirow{2}{*}{2.29B} & \vft{} & $\mathbf{431.1}$ & $241.7$ & $218.9$ & $\mathbf{212.5}$ & $211.0$ & $236.4$ & $250.4$ & $268.0$ \\
 & \ours{} & $439.0$ & $\mathbf{234.9}$ & $\mathbf{213.7}$ & $213.7$ & $\mathbf{196.9}$ & $\mathbf{204.3}^{\dagger}$ & $\mathbf{230.2}^{\dagger}$ & $\mathbf{266.3}$ \\
\midrule
\multicolumn{10}{@{}l}{\textbf{gFID}$\downarrow$} \\
\multirow{2}{*}{49M} & \vft{} & $\mathbf{39.56}^{\dagger}$ & $\mathbf{24.98}^{\dagger}$ & $24.00$ & $23.87$ & $26.12$ & $29.04$ & $35.36$ & $41.13$ \\
 & \ours{} & $42.58$ & $26.12$ & $\mathbf{23.66}$ & $\mathbf{22.37}^{\dagger}$ & $\mathbf{22.27}^{\dagger}$ & $\mathbf{24.05}^{\dagger}$ & $\mathbf{29.04}^{\dagger}$ & $\mathbf{34.30}^{\dagger}$ \\[1pt]
\multirow{2}{*}{85M} & \vft{} & $\mathbf{40.16}^{\dagger}$ & $\mathbf{25.35}$ & $23.43$ & $23.60$ & $25.02$ & $27.70$ & $33.30$ & $38.02$ \\
 & \ours{} & $43.41$ & $25.54$ & $\mathbf{23.32}$ & $\mathbf{21.55}^{\dagger}$ & $\mathbf{21.55}^{\dagger}$ & $\mathbf{22.97}^{\dagger}$ & $\mathbf{27.52}^{\dagger}$ & $\mathbf{32.46}^{\dagger}$ \\[1pt]
\multirow{2}{*}{201M} & \vft{} & $\mathbf{39.49}^{\dagger}$ & $\mathbf{24.35}$ & $22.96$ & $22.52$ & $23.96$ & $26.43$ & $30.77$ & $34.63$ \\
 & \ours{} & $42.61$ & $24.95$ & $\mathbf{22.83}$ & $\mathbf{21.33}^{\dagger}$ & $\mathbf{21.39}^{\dagger}$ & $\mathbf{22.43}^{\dagger}$ & $\mathbf{26.19}^{\dagger}$ & $\mathbf{29.78}^{\dagger}$ \\[1pt]
\multirow{2}{*}{393M} & \vft{} & $\mathbf{39.91}^{\dagger}$ & $24.72$ & $22.51$ & $22.38$ & $23.47$ & $25.95$ & $29.72$ & $33.13$ \\
 & \ours{} & $42.46$ & $\mathbf{24.46}$ & $\mathbf{22.45}$ & $\mathbf{20.90}^{\dagger}$ & $\mathbf{20.68}^{\dagger}$ & $\mathbf{21.67}^{\dagger}$ & $\mathbf{24.72}^{\dagger}$ & $\mathbf{28.05}^{\dagger}$ \\[1pt]
\multirow{2}{*}{679M} & \vft{} & $\mathbf{40.38}^{\dagger}$ & $\mathbf{23.59}^{\dagger}$ & $22.60$ & $22.16$ & $23.03$ & $24.97$ & $28.28$ & $31.51$ \\
 & \ours{} & $42.85$ & $24.56$ & $\mathbf{22.18}$ & $\mathbf{21.43}$ & $\mathbf{20.66}^{\dagger}$ & $\mathbf{21.74}^{\dagger}$ & $\mathbf{24.55}^{\dagger}$ & $\mathbf{27.21}^{\dagger}$ \\[1pt]
\multirow{2}{*}{1.33B} & \vft{} & $\mathbf{40.10}^{\dagger}$ & $\mathbf{24.13}$ & $\mathbf{22.36}$ & $22.06$ & $22.39$ & $24.01$ & $27.72$ & $29.99$ \\
 & \ours{} & $42.97$ & $24.82$ & $22.76$ & $\mathbf{21.05}^{\dagger}$ & $\mathbf{20.81}^{\dagger}$ & $\mathbf{21.46}^{\dagger}$ & $\mathbf{23.93}^{\dagger}$ & $\mathbf{26.54}^{\dagger}$ \\[1pt]
\multirow{2}{*}{2.29B} & \vft{} & $\mathbf{40.57}^{\dagger}$ & $\mathbf{24.35}$ & $22.30$ & $22.07$ & $22.46$ & $24.15$ & $26.71$ & $28.91$ \\
 & \ours{} & $42.72$ & $24.54$ & $\mathbf{22.25}$ & $\mathbf{20.69}^{\dagger}$ & $\mathbf{20.69}^{\dagger}$ & $\mathbf{21.65}^{\dagger}$ & $\mathbf{23.89}^{\dagger}$ & $\mathbf{26.31}^{\dagger}$ \\
\bottomrule
\end{tabular}
\end{table}

\begin{table}[p]
\caption{uCO3D text-to-video generation: semantic alignment. Same samples as \Cref{tab:gen-uco3d}. Class acc.\ is NCM. Bold is better at that $k$ and size; $^\dagger$ marks the better tokenizer where the 95\% paired-bootstrap CI of the difference (single sampling seed) excludes zero. CIs cover every ClipV and class-acc.\ cell; ViCLIP has none.}
\label{tab:gen-uco3d-sem}
\centering
\footnotesize
\setlength{\tabcolsep}{3.2pt}
\begin{tabular}{@{}ll *{8}{c}@{}}
\toprule
AR & & $k{=}1$ & $4$ & $8$ & $16$ & $32$ & $64$ & $128$ & $256$ \\
\midrule
\multicolumn{10}{@{}l}{\textbf{ViCLIP}$\uparrow$} \\
\multirow{2}{*}{49M} & \vft{} & $\mathbf{0.170}$ & $\mathbf{0.210}$ & $\mathbf{0.217}$ & $0.221$ & $0.221$ & $0.220$ & $0.217$ & $0.212$ \\
 & \ours{} & $0.169$ & $0.205$ & $0.214$ & $\mathbf{0.226}$ & $\mathbf{0.230}$ & $\mathbf{0.231}$ & $\mathbf{0.226}$ & $\mathbf{0.220}$ \\[1pt]
\multirow{2}{*}{85M} & \vft{} & $0.170$ & $\mathbf{0.208}$ & $\mathbf{0.218}$ & $0.221$ & $0.222$ & $0.222$ & $0.219$ & $0.216$ \\
 & \ours{} & $0.170$ & $0.206$ & $0.215$ & $\mathbf{0.227}$ & $\mathbf{0.230}$ & $\mathbf{0.232}$ & $\mathbf{0.228}$ & $\mathbf{0.222}$ \\[1pt]
\multirow{2}{*}{201M} & \vft{} & $0.169$ & $\mathbf{0.208}$ & $\mathbf{0.217}$ & $0.222$ & $0.223$ & $0.224$ & $0.221$ & $0.220$ \\
 & \ours{} & $\mathbf{0.170}$ & $0.205$ & $0.215$ & $\mathbf{0.226}$ & $\mathbf{0.229}$ & $\mathbf{0.231}$ & $\mathbf{0.228}$ & $\mathbf{0.225}$ \\[1pt]
\multirow{2}{*}{393M} & \vft{} & $0.168$ & $\mathbf{0.207}$ & $0.216$ & $0.221$ & $0.222$ & $0.223$ & $0.223$ & $0.220$ \\
 & \ours{} & $\mathbf{0.170}$ & $0.206$ & $0.216$ & $\mathbf{0.227}$ & $\mathbf{0.230}$ & $\mathbf{0.232}$ & $\mathbf{0.230}$ & $\mathbf{0.226}$ \\[1pt]
\multirow{2}{*}{679M} & \vft{} & $0.167$ & $\mathbf{0.209}$ & $\mathbf{0.217}$ & $0.222$ & $0.224$ & $0.224$ & $0.225$ & $0.222$ \\
 & \ours{} & $\mathbf{0.168}$ & $0.206$ & $0.214$ & $\mathbf{0.227}$ & $\mathbf{0.230}$ & $\mathbf{0.232}$ & $\mathbf{0.229}$ & $\mathbf{0.227}$ \\[1pt]
\multirow{2}{*}{1.33B} & \vft{} & $0.166$ & $\mathbf{0.208}$ & $0.217$ & $0.221$ & $0.225$ & $0.226$ & $0.225$ & $0.225$ \\
 & \ours{} & $0.166$ & $0.207$ & $0.217$ & $\mathbf{0.226}$ & $\mathbf{0.230}$ & $\mathbf{0.232}$ & $\mathbf{0.230}$ & $\mathbf{0.227}$ \\[1pt]
\multirow{2}{*}{2.29B} & \vft{} & $0.166$ & $\mathbf{0.207}$ & $\mathbf{0.216}$ & $0.221$ & $0.224$ & $0.226$ & $0.226$ & $0.225$ \\
 & \ours{} & $\mathbf{0.167}$ & $0.205$ & $0.215$ & $\mathbf{0.227}$ & $\mathbf{0.230}$ & $\mathbf{0.232}$ & $\mathbf{0.229}$ & $\mathbf{0.228}$ \\
\midrule
\multicolumn{10}{@{}l}{\textbf{ClipV}$\uparrow$} \\
\multirow{2}{*}{49M} & \vft{} & $\mathbf{0.598}^{\dagger}$ & $\mathbf{0.688}$ & $0.703$ & $0.708$ & $0.706$ & $0.700$ & $0.687$ & $0.671$ \\
 & \ours{} & $0.583$ & $0.685$ & $\mathbf{0.711}^{\dagger}$ & $\mathbf{0.731}^{\dagger}$ & $\mathbf{0.736}^{\dagger}$ & $\mathbf{0.736}^{\dagger}$ & $\mathbf{0.719}^{\dagger}$ & $\mathbf{0.701}^{\dagger}$ \\[1pt]
\multirow{2}{*}{85M} & \vft{} & $\mathbf{0.599}^{\dagger}$ & $0.688$ & $0.707$ & $0.710$ & $0.712$ & $0.707$ & $0.694$ & $0.681$ \\
 & \ours{} & $0.584$ & $\mathbf{0.689}$ & $\mathbf{0.713}^{\dagger}$ & $\mathbf{0.735}^{\dagger}$ & $\mathbf{0.739}^{\dagger}$ & $\mathbf{0.739}^{\dagger}$ & $\mathbf{0.725}^{\dagger}$ & $\mathbf{0.710}^{\dagger}$ \\[1pt]
\multirow{2}{*}{201M} & \vft{} & $\mathbf{0.597}^{\dagger}$ & $0.690$ & $0.709$ & $0.715$ & $0.717$ & $0.716$ & $0.705$ & $0.694$ \\
 & \ours{} & $0.584$ & $\mathbf{0.692}$ & $\mathbf{0.713}^{\dagger}$ & $\mathbf{0.738}^{\dagger}$ & $\mathbf{0.743}^{\dagger}$ & $\mathbf{0.742}^{\dagger}$ & $\mathbf{0.731}^{\dagger}$ & $\mathbf{0.721}^{\dagger}$ \\[1pt]
\multirow{2}{*}{393M} & \vft{} & $\mathbf{0.594}^{\dagger}$ & $0.687$ & $0.708$ & $0.716$ & $0.716$ & $0.717$ & $0.709$ & $0.702$ \\
 & \ours{} & $0.585$ & $\mathbf{0.693}^{\dagger}$ & $\mathbf{0.720}^{\dagger}$ & $\mathbf{0.740}^{\dagger}$ & $\mathbf{0.744}^{\dagger}$ & $\mathbf{0.745}^{\dagger}$ & $\mathbf{0.737}^{\dagger}$ & $\mathbf{0.726}^{\dagger}$ \\[1pt]
\multirow{2}{*}{679M} & \vft{} & $\mathbf{0.592}^{\dagger}$ & $0.689$ & $0.709$ & $0.717$ & $0.719$ & $0.719$ & $0.716$ & $0.704$ \\
 & \ours{} & $0.582$ & $\mathbf{0.694}^{\dagger}$ & $\mathbf{0.720}^{\dagger}$ & $\mathbf{0.742}^{\dagger}$ & $\mathbf{0.748}^{\dagger}$ & $\mathbf{0.749}^{\dagger}$ & $\mathbf{0.739}^{\dagger}$ & $\mathbf{0.732}^{\dagger}$ \\[1pt]
\multirow{2}{*}{1.33B} & \vft{} & $\mathbf{0.590}^{\dagger}$ & $0.690$ & $0.708$ & $0.716$ & $0.722$ & $0.722$ & $0.719$ & $0.713$ \\
 & \ours{} & $0.579$ & $\mathbf{0.697}^{\dagger}$ & $\mathbf{0.722}^{\dagger}$ & $\mathbf{0.740}^{\dagger}$ & $\mathbf{0.747}^{\dagger}$ & $\mathbf{0.749}^{\dagger}$ & $\mathbf{0.742}^{\dagger}$ & $\mathbf{0.731}^{\dagger}$ \\[1pt]
\multirow{2}{*}{2.29B} & \vft{} & $\mathbf{0.590}^{\dagger}$ & $0.687$ & $0.709$ & $0.716$ & $0.721$ & $0.724$ & $0.720$ & $0.715$ \\
 & \ours{} & $0.580$ & $\mathbf{0.695}^{\dagger}$ & $\mathbf{0.721}^{\dagger}$ & $\mathbf{0.741}^{\dagger}$ & $\mathbf{0.749}^{\dagger}$ & $\mathbf{0.750}^{\dagger}$ & $\mathbf{0.742}^{\dagger}$ & $\mathbf{0.735}^{\dagger}$ \\
\midrule
\multicolumn{10}{@{}l}{\textbf{class acc.}$\uparrow$} \\
\multirow{2}{*}{49M} & \vft{} & $0.101$ & $0.219$ & $0.246$ & $0.249$ & $0.247$ & $0.231$ & $0.203$ & $0.187$ \\
 & \ours{} & $\mathbf{0.102}$ & $\mathbf{0.240}^{\dagger}$ & $\mathbf{0.290}^{\dagger}$ & $\mathbf{0.309}^{\dagger}$ & $\mathbf{0.316}^{\dagger}$ & $\mathbf{0.298}^{\dagger}$ & $\mathbf{0.255}^{\dagger}$ & $\mathbf{0.220}^{\dagger}$ \\[1pt]
\multirow{2}{*}{85M} & \vft{} & $0.103$ & $0.223$ & $0.245$ & $0.255$ & $0.250$ & $0.239$ & $0.221$ & $0.202$ \\
 & \ours{} & $\mathbf{0.106}$ & $\mathbf{0.254}^{\dagger}$ & $\mathbf{0.290}^{\dagger}$ & $\mathbf{0.315}^{\dagger}$ & $\mathbf{0.317}^{\dagger}$ & $\mathbf{0.306}^{\dagger}$ & $\mathbf{0.267}^{\dagger}$ & $\mathbf{0.241}^{\dagger}$ \\[1pt]
\multirow{2}{*}{201M} & \vft{} & $0.101$ & $0.221$ & $0.251$ & $0.259$ & $0.265$ & $0.259$ & $0.237$ & $0.225$ \\
 & \ours{} & $\mathbf{0.103}$ & $\mathbf{0.259}^{\dagger}$ & $\mathbf{0.296}^{\dagger}$ & $\mathbf{0.322}^{\dagger}$ & $\mathbf{0.330}^{\dagger}$ & $\mathbf{0.315}^{\dagger}$ & $\mathbf{0.279}^{\dagger}$ & $\mathbf{0.255}^{\dagger}$ \\[1pt]
\multirow{2}{*}{393M} & \vft{} & $0.099$ & $0.220$ & $0.260$ & $0.263$ & $0.264$ & $0.260$ & $0.249$ & $0.236$ \\
 & \ours{} & $\mathbf{0.105}$ & $\mathbf{0.261}^{\dagger}$ & $\mathbf{0.306}^{\dagger}$ & $\mathbf{0.343}^{\dagger}$ & $\mathbf{0.336}^{\dagger}$ & $\mathbf{0.324}^{\dagger}$ & $\mathbf{0.304}^{\dagger}$ & $\mathbf{0.273}^{\dagger}$ \\[1pt]
\multirow{2}{*}{679M} & \vft{} & $0.097$ & $0.223$ & $0.258$ & $0.269$ & $0.279$ & $0.270$ & $0.261$ & $0.244$ \\
 & \ours{} & $\mathbf{0.099}$ & $\mathbf{0.258}^{\dagger}$ & $\mathbf{0.299}^{\dagger}$ & $\mathbf{0.334}^{\dagger}$ & $\mathbf{0.340}^{\dagger}$ & $\mathbf{0.332}^{\dagger}$ & $\mathbf{0.302}^{\dagger}$ & $\mathbf{0.276}^{\dagger}$ \\[1pt]
\multirow{2}{*}{1.33B} & \vft{} & $0.092$ & $0.222$ & $0.254$ & $0.269$ & $0.275$ & $0.273$ & $0.265$ & $0.255$ \\
 & \ours{} & $\mathbf{0.102}^{\dagger}$ & $\mathbf{0.260}^{\dagger}$ & $\mathbf{0.305}^{\dagger}$ & $\mathbf{0.330}^{\dagger}$ & $\mathbf{0.344}^{\dagger}$ & $\mathbf{0.333}^{\dagger}$ & $\mathbf{0.314}^{\dagger}$ & $\mathbf{0.289}^{\dagger}$ \\[1pt]
\multirow{2}{*}{2.29B} & \vft{} & $0.092$ & $0.220$ & $0.263$ & $0.280$ & $0.288$ & $0.286$ & $0.282$ & $0.270$ \\
 & \ours{} & $\mathbf{0.100}^{\dagger}$ & $\mathbf{0.261}^{\dagger}$ & $\mathbf{0.312}^{\dagger}$ & $\mathbf{0.337}^{\dagger}$ & $\mathbf{0.352}^{\dagger}$ & $\mathbf{0.337}^{\dagger}$ & $\mathbf{0.321}^{\dagger}$ & $\mathbf{0.302}^{\dagger}$ \\
\bottomrule
\end{tabular}
\end{table}

\begin{table}[p]
\caption{Decoder-REPA readout versus $k$ and decoder noise $\sigma$. Cosine between DINOv2 features of the first frame of each latent group and the decoder-REPA readout from the first $k$ tokens per frame, 256 validation clips. $\sigma{=}1$ is pure noise, as in generation. Bold is better at that $k$ and $\sigma$.}
\label{tab:repa}
\centering
\footnotesize
\setlength{\tabcolsep}{3.2pt}
\begin{tabular}{@{}ll *{8}{c}@{}}
\toprule
$\sigma$ & & $k{=}1$ & $4$ & $8$ & $16$ & $32$ & $64$ & $128$ & $256$ \\
\midrule
\multicolumn{10}{@{}l}{\textbf{Kinetics-600}} \\
\multirow{2}{*}{$1$} & \vft{} & $0.480$ & $0.529$ & $0.555$ & $0.581$ & $0.599$ & $0.614$ & $0.624$ & $0.631$ \\
 & \ours{} & $\mathbf{0.491}$ & $\mathbf{0.565}$ & $\mathbf{0.593}$ & $\mathbf{0.623}$ & $\mathbf{0.646}$ & $\mathbf{0.668}$ & $\mathbf{0.688}$ & $\mathbf{0.707}$ \\[1pt]
\multirow{2}{*}{$0.75$} & \vft{} & $0.517$ & $0.552$ & $0.571$ & $0.593$ & $0.608$ & $0.621$ & $0.630$ & $0.636$ \\
 & \ours{} & $\mathbf{0.527}$ & $\mathbf{0.584}$ & $\mathbf{0.608}$ & $\mathbf{0.634}$ & $\mathbf{0.654}$ & $\mathbf{0.673}$ & $\mathbf{0.692}$ & $\mathbf{0.709}$ \\[1pt]
\multirow{2}{*}{$0.5$} & \vft{} & $\mathbf{0.612}$ & $0.624$ & $0.630$ & $0.639$ & $0.645$ & $0.650$ & $0.654$ & $0.656$ \\
 & \ours{} & $0.605$ & $\mathbf{0.635}$ & $\mathbf{0.650}$ & $\mathbf{0.665}$ & $\mathbf{0.677}$ & $\mathbf{0.689}$ & $\mathbf{0.702}$ & $\mathbf{0.716}$ \\[1pt]
\multirow{2}{*}{$0.25$} & \vft{} & $\mathbf{0.674}$ & $\mathbf{0.678}$ & $\mathbf{0.679}$ & $0.682$ & $0.683$ & $0.684$ & $0.685$ & $0.685$ \\
 & \ours{} & $0.648$ & $0.667$ & $0.678$ & $\mathbf{0.688}$ & $\mathbf{0.696}$ & $\mathbf{0.704}$ & $\mathbf{0.712}$ & $\mathbf{0.722}$ \\
\midrule
\multicolumn{10}{@{}l}{\textbf{uCO3D}} \\
\multirow{2}{*}{$1$} & \vft{} & $0.548$ & $0.607$ & $0.629$ & $0.641$ & $0.651$ & $0.659$ & $0.671$ & $0.683$ \\
 & \ours{} & $\mathbf{0.566}$ & $\mathbf{0.659}$ & $\mathbf{0.685}$ & $\mathbf{0.705}$ & $\mathbf{0.721}$ & $\mathbf{0.733}$ & $\mathbf{0.744}$ & $\mathbf{0.752}$ \\[1pt]
\multirow{2}{*}{$0.75$} & \vft{} & $0.589$ & $0.626$ & $0.641$ & $0.652$ & $0.659$ & $0.666$ & $0.677$ & $0.688$ \\
 & \ours{} & $\mathbf{0.604}$ & $\mathbf{0.673}$ & $\mathbf{0.694}$ & $\mathbf{0.711}$ & $\mathbf{0.724}$ & $\mathbf{0.735}$ & $\mathbf{0.746}$ & $\mathbf{0.753}$ \\[1pt]
\multirow{2}{*}{$0.5$} & \vft{} & $\mathbf{0.672}$ & $0.678$ & $0.682$ & $0.685$ & $0.687$ & $0.690$ & $0.695$ & $0.701$ \\
 & \ours{} & $0.671$ & $\mathbf{0.699}$ & $\mathbf{0.712}$ & $\mathbf{0.722}$ & $\mathbf{0.731}$ & $\mathbf{0.738}$ & $\mathbf{0.747}$ & $\mathbf{0.754}$ \\[1pt]
\multirow{2}{*}{$0.25$} & \vft{} & $\mathbf{0.716}$ & $\mathbf{0.715}$ & $0.716$ & $0.716$ & $0.716$ & $0.717$ & $0.718$ & $0.719$ \\
 & \ours{} & $0.697$ & $0.714$ & $\mathbf{0.723}$ & $\mathbf{0.731}$ & $\mathbf{0.737}$ & $\mathbf{0.742}$ & $\mathbf{0.749}$ & $\mathbf{0.755}$ \\
\bottomrule
\end{tabular}
\end{table}

\end{document}